\documentclass[10.5pt,onecolumn]{article}

\usepackage[a4paper,total={6in, 8in}]{geometry}

\PassOptionsToPackage{numbers, sort&compress}{natbib}

\usepackage[utf8]{inputenc} 
\usepackage[T1]{fontenc}    
\usepackage{hyperref}       
\usepackage{url}  

\PassOptionsToPackage{bookmarks=false}{hyperref}

\usepackage[depth=-1]{bookmark}

\usepackage{booktabs}       
\usepackage{amsfonts}       
\usepackage{nicefrac}       
\usepackage{microtype}      
\usepackage{xcolor}         
\usepackage{mathrsfs} 
\usepackage{amsmath}
\usepackage{amsthm}
\usepackage{graphicx} 
\usepackage{subfig}    
\usepackage{float} 
\usepackage{algorithm}
\usepackage{algorithmicx}
\usepackage{algpseudocode}
\usepackage{booktabs}
\usepackage{multirow}
\usepackage{setspace}

\usepackage{wrapfig}

\usepackage{accents}
\makeatletter
\def\wid{\check{{\cc@style\underline{\mskip9.5mu}}}}
\def\Wideubar{\underaccent{{\cc@style\underline{\mskip6mu}}}}
\makeatother

\makeatletter
\def\wideubar{\underaccent{{\cc@style\underline{\mskip9.5mu}}}}
\def\Wideubar{\underaccent{{\cc@style\underline{\mskip6mu}}}}
\makeatother

\makeatletter
\def\widebar{\accentset{{\cc@style\underline{\mskip9.5mu}}}}
\def\Widebar{\accentset{{\cc@style\underline{\mskip6mu}}}}
\makeatother

\theoremstyle{remark}

\allowdisplaybreaks

\makeatletter
\newcommand*\bigcdot{\mathpalette\bigcdot@{.5}}
\newcommand*\bigcdot@[2]{\mathbin{\vcenter{\hbox{\scalebox{#2}{$\m@th#1\bullet$}}}}}
\makeatother

\title{\textbf{Soft Decomposed Policy-Critic: Bridging the Gap for Effective Continuous Control with Discrete RL}}

\begin{document}
	
\author{Yechen Zhang, Jian Sun, Gang Wang, Zhuo Li, Wei Chen}

	\maketitle
	
\begin{abstract}
	Discrete reinforcement learning (RL) algorithms have demonstrated exceptional performance in solving sequential decision tasks with discrete action spaces, such as Atari games. However, their effectiveness is hindered when applied to continuous control problems due to the challenge of dimensional explosion. In this paper, we present the Soft Decomposed Policy-Critic (SDPC) architecture, which combines soft RL and actor-critic techniques with discrete RL methods to overcome this limitation. SDPC discretizes each action dimension independently and employs a shared critic network to maximize the soft $Q$-function. This novel approach enables SDPC to support two types of policies: decomposed actors that lead to the Soft Decomposed Actor-Critic (SDAC) algorithm, and decomposed $Q$-networks that generate Boltzmann soft exploration policies, resulting in the Soft Decomposed-Critic Q (SDCQ) algorithm. Through extensive experiments, we demonstrate that our proposed approach outperforms state-of-the-art continuous RL algorithms in a variety of continuous control tasks, including Mujoco's Humanoid and Box2d's BipedalWalker. These empirical results validate the effectiveness of the SDPC architecture in addressing the challenges associated with continuous control.

\end{abstract}

\section{Introduction}

Reinforcement learning (RL) \cite{sutton2018reinforcement} is a powerful class of machine learning methods that enables models to learn and make decisions through interactions with the environment. RL can be divided into discrete RL, where the action space consists of a set of discrete actions, and continuous RL, where the action space is continuous and potentially infinite. While continuous RL algorithms, particularly those employing the actor-critic structure \cite{konda1999actor}, have been widely used for continuous control tasks to avoid the dimensional explosion problem associated with discretizing the action space, they often exhibit fragility and sensitivity to hyperparameters, leading to training failures in certain scenarios.

To overcome these limitations and leverage the benefits of discrete RL in continuous control tasks, researchers have explored the use of discrete RL algorithms. One approach to address the dimensional explosion problem caused by action discretization is to discretize each action dimension independently in the high-dimensional continuous action space, thereby avoiding exponential growth in the number of discrete actions. Existing methods, such as branching dueling $Q$-network (BDQ) \cite{tavakoli2018action}, have attempted to tackle this problem using dimensional-independent discretization. However, evaluations of these methods have shown drawbacks in terms of training efficiency and final performance.

In this work, we propose a novel architecture termed soft decomposed policy-critic (SDPC) to effectively solve continuous control problems using discrete RL algorithms. The term ``decomposed'' in SDPC refers to the separate discretization of continuous actions on each action dimension. While the ``decomposed policy network'' effectively addresses the dimensional explosion problem, SDPC takes a step further by incorporating the actor-critic framework and soft RL. This enables a continuous critic that maximizes the soft $Q$-value function and provides evaluations for all discrete actions in each action dimension simultaneously. One key advantage of SDPC is its flexibility in performing offline value iterations, significantly improving training efficiency through experience replay. To validate our claims, we establish a decomposed discrete Markov decision process (MDP) model to demonstrate the feasibility of experience replay in SDPC, while also highlighting the drawbacks of offline iteration methods like BDQ.

SDPC supports two types of policy networks: the decomposed actor network and the decomposed $Q$-network. The former generates dimensional-independent discrete action distributions, which can easily integrate with soft RL to offer stochastic policies on each action dimension. Our proposed algorithm called soft decomposed actor-critic (SDAC), builds upon the SDPC architecture and leverages the decomposed actor network. Through extensive evaluation in diverse benchmark environments, we show that SDAC outperforms or achieves comparable performance to the continuous Soft Actor-Critic (SAC) algorithm \cite{haarnoja2018soft1, haarnoja2018soft2}, thereby demonstrating the efficiency of our SDPC structure.

While the decomposed $Q$-network cannot directly associate with soft RL, we introduce Boltzmann policies to generate stochastic action distributions. We analyze the equivalence between the stochastic policies of the decomposed actor and the Boltzmann policies of the decomposed $Q$-networks, which leads to the development of our Soft Decomposed-Critic Q (SDCQ) algorithm. SDCQ is a value-based RL algorithm that fits discrete $Q$ values by minimizing the error between the decomposed $Q$-network and the critic network. Furthermore, it directly controls the exploration rate through adaptive temperature. Our implementation of SDCQ, incorporating double $Q$-networks and multi-step temporal-difference (TD) learning, demonstrates remarkable improvements over continuous actor-critic methods in substantial numerical tests across various continuous control tasks. SDCQ converges with fewer training steps, typically less than half of those required by SAC, and achieves the highest final performance.

  \section{Related Works}

  Discrete reinforcement learning (RL) algorithms, particularly value-based methods, have demonstrated exceptional performance in sequential decision-making tasks. Tabular methods, such as $Q$-learning \cite{watkins1992q}, excel in handling discrete state and action spaces. However, they face challenges when dealing with complex, continuous state information. To address this issue, the introduction of neural networks in deep $Q$-networks (DQN) \cite{mnih2015human} enabled the algorithm to solve a wide range of complicated tasks, including stock market forecasting \cite{carta2021multi}, gesture recognition \cite{valdivieso2023recognition}, and path planning \cite{li2022multirobot}. Strategies like double $Q$-networks and prioritized experience replay further enhanced DQN's performance, surpassing human-level scores. Nevertheless, discrete RL algorithms still struggle when applied to continuous control tasks with multi-dimensional continuous action spaces due to the dimensional explosion problem that arises during action discretization. With $M$ action dimensions and each dimension discretized into $N$ slices, the algorithm would need to handle an exponentially growing number of discrete actions, reaching $M^{N}$ possibilities.

 To tackle continuous control tasks, researchers have turned their attention to actor-critic algorithms. While on-policy methods like trust region policy optimization \cite{schulman2015trust, wu2017scalable} and proximal policy optimization \cite{schulman2017proximal} improve stability, they often suffer from poor sample efficiency. Consequently, off-policy actor-critic methods have become the most effective approach for continuous control tasks since the proposal of deterministic policy gradients \cite{silver2014deterministic}. These methods can be broadly categorized into two types: deterministic algorithms and stochastic algorithms. Deterministic algorithms, such as the deep deterministic policy gradient (DDPG) \cite{lillicrap2015continuous}, distributed distributional deterministic policy gradients (D4PG) \cite{barth2018distributed}, and twin delayed DDPG (TD3) \cite{fujimoto2018addressing}, are relatively easier to implement. On the other hand, stochastic algorithms, particularly soft actor-critic (SAC) \cite{haarnoja2018soft2}, offer higher training efficiency.

   The outstanding performance of SAC can be attributed to its utilization of soft RL, which maximizes both value functions and the entropy of stochastic policies. The framework of maximum entropy has gained popularity in various RL research areas, including guided policy search \cite{levine2016end}, $Q$-learning methods \cite{haarnoja2017reinforcement}, and inverse RL \cite{aghasadeghi2011maximum}. Among these approaches, SAC stands out as the most effective method for introducing maximum entropy to the actor-critic architecture. In SAC, a critic network is employed to fit a soft $Q$-function, while the policy network outputs Gaussian action distributions. The soft $Q$-function is used to update the mean and variance of the Gaussian distributions, significantly improving the robustness and exploration of SAC. The second version of SAC introduced an adaptive temperature adjustment to constrain the entropy of stochastic policies to a predefined target entropy, simplifying the tuning process and leading to further improvements during evaluations.
  
  Despite the notable successes achieved by continuous actor-critic methods, they are not exempt from limitations. In particular, the SAC algorithm may encounter challenges when applied to the MuJoCo benchmark task `Reacher-v2'. As a result, researchers have sought modifications to discrete RL methods in order to address the difficulties posed by continuous control problems. For example, Perrusquia et al. \cite{perrusquia2021multi} focused on robot control, while Kurniawan et al. \cite{kurniawan2022discrete} investigated Gym benchmark environments.  
  A highly effective approach for handling multi-dimensional continuous actions is to discretize each dimension independently, resulting in a total of $M \times N$ discrete actions. Several attempts have been made following this approach. Tavakoli et al. \cite{tavakoli2018action} proposed a branching dueling network architecture that incorporates a centralized value function and independent $Q$ functions for each action dimension. Metz et al. \cite{metz2017discrete} introduced a next-step prediction model to sequentially learn the discrete $Q$-value for each dimension. Other relevant works in this area include those by Tang et al. \cite{tang2020discretizing}, Jaskowski et al. \cite{jaskowski2018reinforcement}, and Andrychowicz et al. \cite{andrychowicz2020learning}. However, it is important to note that these methods are typically less effective than offline actor-critic algorithms.

\section{Soft decomposed policy-critic: Continuous soft \emph{Q}-function for evaluating discrete actions}\label{SDAC}

In this section, we present the architecture of SDPC and its corresponding neural networks (NN). We analyze the value functions of discrete policies, establish a decomposed discrete Markov decision process (MDP), and bridge it to a continuous MDP. This bridge between the two MDPs highlights the effectiveness of SDPC's policy iterations. Further, we introduce the basic SDAC algorithm, which leverages continuous soft $Q$-functions to optimize the distributions of discrete actions.

\subsection{Architecture of Soft Decomposed Policy-Critic}\label{sec:arch}
The SDPC architecture consists of decomposed discrete policies and continuous critics. We start constructing SDPC with decomposed discretization, where a continuous action space $\mathcal{A} := {a \in \mathbb{R}^M}$ with $M$ dimensions is decomposed into individual continuous actions and discretized into independent discrete action spaces. Without loss of generality, we assume that each dimension of the continuous action $a(t)=[a_1(t) \,a_2(t)\, \cdots \,a_M(t)]^{\sf T}$ is constrained to $a_m(t)  \in [-1, 1]$ for $m\in[ M] := {1,2,\ldots,M}$, allowing us to discretize each continuous action dimension $m$ into $N$ fractions.

The resulting discrete action space $\mathcal{A}^d_m$ for the $m$-th dimension is denoted as
\begin{equation} \label{actionspace}
	\mathcal{A}_{m}^d:=\left\{a^d_{m,1},\, a^d_{m,2},\,...,\, a^d_{m,N}\right\}, \quad {\rm for}\quad  m\in[ M], 
\end{equation}
where $a^d_{m,n}$ represents the $n$-th discrete action corresponding to the $m$-th action dimension. In SDPC, we construct an independent discrete policy for each dimension $m$, covering all the discrete actions in $\mathcal{A}^d_{m}$. This results in a total of $M \times N$ discrete actions across all $M$ action dimensions, facilitating the use of deep neural networks (DNNs).

\begin{figure}
	\centering
	\includegraphics[width=5in]{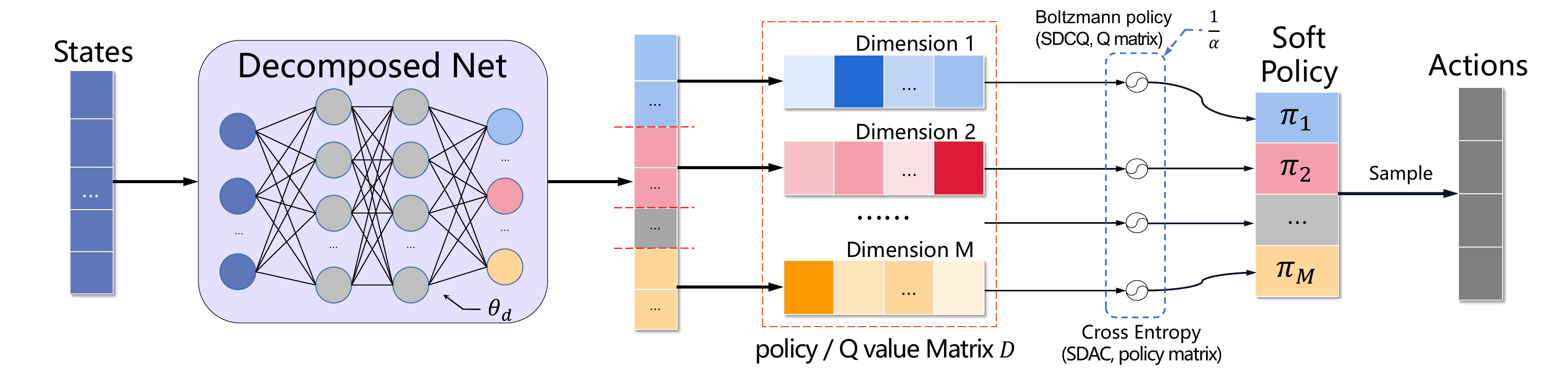}
	\caption{Structure of the decomposed policy network. The matrix $D$ can be used to generate policies straightly through cross entropy in SDAC, or learn discrete Q values for Boltzmann policy in SDCQ.}
	\label{img1}
\end{figure}

For the corresponding discrete policy network, previous results following the separate-discretization routine have proposed feasible constructions for decomposed discrete action spaces. Examples include dimensional independent dueling $Q$-networks (IDQ) \cite{tavakoli2018action} and network branches for action dimensions in BDQ \cite{tavakoli2018action}. In contrast, this present work designs a decomposed policy network, which is more effective. Our approach first outputs an array through multi-layer perceptrons (MLPs) and then reconstructs it into an $M \times N$ matrix, where each row of the matrix corresponds to a single-dimensional discrete policy. We denote this matrix as $D(s(t);\theta_d)$, where $s(t)$ represents the state and $\theta_d$ denotes the parameters of the MLPs. We use $d_{m,n}(s(t);\theta_d)$ to indicate the $n$-th element of the $m$-th row in $D(s(t);\theta_d)$. Figure \ref{img1} illustrates the structure of the decomposed policy network.

Within the SDPC architecture, two types of policies are designed: the decomposed actor and the decomposed Boltzmann policy. In this section, we focus on the decomposed actor, which generates stochastic policies through cross-entropy on each row of the matrix $D$. The discrete policies can be denoted as follows
 \begin{equation}\label{policy}
 	\pi(a(t)|s(t);\theta_d) = \{\pi_m\}_{m\in [M]}, \ \ {\rm with} \ \  \pi_m(a^d_{m,n}|s(t);\theta_d)=\frac{{\rm exp}(d_{m,n}(s(t);\theta_d))}{\sum^N_{j=1}{\rm exp}(d_{m,j}(s(t);\theta_d))}.
 \end{equation}
 Here, $\pi(a(t)|s(t);\theta_d)$ denotes the discrete policy conditioned on the state $s(t)$, and $\pi_m(a^d_{m,n}|s(t);\theta_d)$ denotes the probability of selecting the $n$-th discrete action in the $m$-th action dimension.

 The key contribution of our SDPC architecture lies in the continuous critic networks, which provide evaluations for the discrete actions. The critics in SDPC have the same structure as those in TD3. They take the state $s(t)$ and multi-dimensional continuous action $a(t):=\{a_1(t),a_2(t),  \cdots, a_M(t)\}$ as input and output the corresponding soft $Q$-function $Q(s(t), a(t); \theta_Q)$, where $\theta_Q$ represents the parameters of the critic network. We optimize our critic network using soft value iterations, by minimizing the following loss function
\begin{equation}\label{valueits}
	J^Q (\theta_Q)= \Big[Q(s(t), a(t); \theta_{Q})- \Big(r_t +\gamma \big(\alpha \mathcal{H}(s(t+1);\theta_d)+Q(s(t+1), \widetilde{a}(t+1); \theta_{Q}')\big)\Big) \Big]^2,
\end{equation}
where $r_t=r(s(t), a(t))$ represents the reward obtained by executing action $a(t)$ in state $s(t)$, $\gamma$ is the discount factor, $\alpha$ is the adaptive temperature, $\theta_Q'$ denotes the parameters of the target critic network, and $\widetilde{a}(t+1)$ is a target action sampled from the discrete policy $\pi$ in Equation \eqref{policy}.

Unlike continuous methods, the entropy $\mathcal{H}$ of the decomposed discrete policies in SDPC can be calculated directly without the need for sampling. We can compute the entropy $\mathcal{H}_m$ for each action dimension individually and then sum them to obtain the global entropy $\mathcal{H}$:
\begin{equation}\label{entropy}
	\mathcal{H}=\sum_{i=1}^M \mathcal{H}_m, \ \ {\rm where} \ \  \mathcal{H}_m(s(t);\theta_d)=\sum_{j=1}^N \pi_m(a^d_{m,n}|s(t);\theta_d) \ {\rm log} \ \pi_m(a^d_{m,n}|s(t);\theta_d).
\end{equation}   
The high efficiency of our SDPC architecture is achieved through policy iterations, enabled by the continuous soft $Q$-function and the entropy function. To explain the policy iteration process between the decomposed actors and the continuous critic, we establish a decomposed discrete MDP to represent the discrete policies. The policy loss function naturally arises from the connection between the discrete MDP and the continuous MDP.

\subsection{Bridging the gap between discrete MDP and continuous MDP}\label{MDPbridge}

We begin by introducing the original continuous control MDP denoted as $\{\mathcal{S}, \mathcal{A}, p, r\}$. Here, $\mathcal{S}$ represents the state space, $p(s(t+1)|s(t), a(t))$ represents the transition probability, and $r(s(t), a(t))$ is the reward function. In soft RL, the goal is to find an optimal policy that maximizes the value function, which is the expected sum of discounted rewards and entropy. The $Q$-value function based on the continuous MDP is defined as follows:
\begin{equation}\label{continuousQ}
	Q(s(t), a(t))=r(s(t), a(t)) + \mathbb{E}_{s(\!t\!+\!k\!)\!\sim\! p, a(\!t\!+\!k\!)\!\sim\! \pi}\!\Big[\sum_{k=1}^{\infty}\gamma^k \big(r(s(t\!+\!k), a(t\!+\!k)) + \alpha\mathcal{H}(s(t\!+\!k))\big)\Big].
\end{equation}  
Here, $Q(s(t), a(t))$ represents the value of taking action $a(t)$ in state $s(t)$. The value function includes the immediate reward $r(s(t), a(t))$ and the expected discounted future rewards and entropy.

Considering the decomposition of the global action space $\mathcal{A}$ into single-dimensional sub-action spaces $\mathcal{A}^d_m$ in our SDPC architecture, we introduce a decomposed MDP $\{\mathcal{S}, \mathcal{A}^d_m, p_m, r_m\}$ to capture the independent discrete policy iterations on each action dimension. Here, we define an ``exclusive policy'' $\overline{\pi}_m=\{\pi_i| {i\in [M], i\neq m} \}$ as the set of discrete policies in each action dimension except for $m$. In the decomposed MDP corresponding to dimension $m$, $\overline{\pi}_m$ remains fixed in the transition probability $p_m$ and the reward function $r_m$, yielding the following formulation
\begin{subequations}
	\label{MDPrelations}
	\begin{align}
		&p_m\big(s(t+1)|s(t), a_m(t);\overline{\pi}_m\big) =  \ p\big( s(t+1)|s(t), \{a_m(t), \overline{a}_m(t)\} \big) \\ \label{ptransform}
		&r_m\big(s(t), a_m(t);\overline{\pi}_m\big) = \ r\big(s(t), \{a_m(t), \overline{a}_m(t)\} \big) + \alpha\overline{\mathcal{H}}_m\big(s(t), \overline{\pi}_m\big)\\[-0.1cm] \label{rtransform}
		&{\rm where} \quad \overline{a}_m(t)\sim\overline{\pi}_m\big(\overline{a}_m(t) | s(t)\big),\quad \overline{\mathcal{H}}_m\big(s(t), \overline{\pi}_m\big) = \sum_{i=1, i\neq m}^{M}\mathcal{H}_i\notag.
	\end{align}
\end{subequations}
In the decomposed MDP, we can construct an $m$-dimensional integral action with an arbitrary $a_m(t)\in \mathcal{A}^d_m$, while the ``exclusive action'' $\overline{a}_m(t)$ is sampled from $\overline{\pi}_m$. The aim of our decomposed MDP is to optimize $\overline{\pi}_m$ in order to maximize the discrete soft value function. The corresponding $Q$ function is defined as follows
\begin{align}
	Q_d(s(t), a_m(t); &\overline{\pi}_m) = r_m(s(t),a_m(t);\overline{\pi}_m) +  \\[-0.05cm]
	&\mathbb{E}_{s(\!t\!+\!k\!)\sim p_m, a_m(\!t\!+\!k\!)\sim \pi_m}\Big[\sum_{k=1}^{\infty}\gamma^k \big(r_m(s(t+k), a_m(t+k)) + \alpha\mathcal{H}_m(s(t+k))\big)\Big]\notag .
\end{align}
It is worth noting that fitting $Q_d$ through offline experience replay poses a challenge. The dynamic nature of $p_m$ and $r_m$ during policy optimization leads to offline experience samples ${s(t), a(t), r_t, s(t+1)}$ that do not align with the current decomposed MDPs, resulting in biased off-policy TD targets. This observation highlights the limited efficiency of prior off-policy methods with single-dimensional discretization. Existing methods perform offline value iterations based on a simplification of the decomposed MDPs with $\alpha=0$. To address this issue, a practical approach for integrating decomposed policies with offline training involves fitting a continuous Q function first using Equation \eqref{valueits}, and then optimizing the decomposed policies using the following bridge between the discrete and continuous $Q$ functions
\begin{equation}\label{Q-bridge}
	\mathbb{E}_{a_m(t)\sim\pi_m}\big[Q_d(s(t), a_m(t);\overline{\pi}_m)\big] = \mathbb{E}_{a(t)\sim\pi}\big[Q(s(t), a(t))\big].
\end{equation}  
We provide a proof for Equation \eqref{Q-bridge} in Appendix \ref{appc1}. With the bridge established, continuous $Q$ functions can be used to construct losses for discrete policy optimization. We define the soft policy loss functions for each action dimension of SDPC as the following KL divergence
\begin{align}
	\label{KLloss}
	J^{\pi}_m(\theta_{d}) 
	&={\rm D_{KL}} \bigg(\pi_m(\bigcdot 
	\ |s(t);\theta_{d})\Bigg|\Bigg|\frac{\exp\big(\frac{1}{\alpha}Q_d(s(t),\bigcdot \ ;
		\overline{\pi}_m)\big)}{\sum_{j=1}^{N}\exp\big(\frac{1}{\alpha}Q_d(s(t),a^d_{m,j} ;
		\overline{\pi}_m)\big)} \bigg)	\\
	&={\rm D_{KL}} \bigg(\pi_m(\bigcdot 
	\ |s(t);\theta_{d})\Bigg|\Bigg|\frac{\exp\big(\frac{1}{\alpha}Q(s(t),\{\bigcdot \ ,\overline{a}_m(t)\};
		\theta_Q)\big)}{\sum_{j=1}^{N}\exp\big(\frac{1}{\alpha}Q(s(t),\{a^d_{m,j} ,\overline{a}_m(t)\};
		\theta_Q)\big)} \bigg) \notag
\end{align}  
In a single iteration, we can simultaneously compute $J^{\pi}_m(\theta_{d})$ for all $M$ action dimensions and then integrate them into a global discrete policy loss function $J^{\pi}(\theta_{d})=\frac{1}{M}\sum_{m = 1}^{M} J^{\pi}_m(\theta_{d})$.

\subsection{Soft decomposed actor-critic}\label{practicalSDAC}

In this section, we present our soft decomposed actor-critic (SDAC) algorithm, which builds upon the extension of SAC to discrete policies by Christodoulou \cite{christodoulou2019soft}. SDAC is specifically designed to tackle continuous control problems and comprises a critic network, a target critic network, and a decomposed actor network. The critic iteratively updates its soft $Q$-function using Equation \eqref{valueits}, while the decomposed discrete actor minimizes the policy loss function defined in Equation \eqref{KLloss}. During interactions with the environment, SDAC employs discrete policies and stores transition experiences $\{s(t),a(t), r_t, s(t+1)\}$ to facilitate soft $Q$-value iteration in the critic. We introduce several practical modifications to enhance the implementation of SDAC.

\begin{figure}
	\centering
	\includegraphics[width=5.5in]{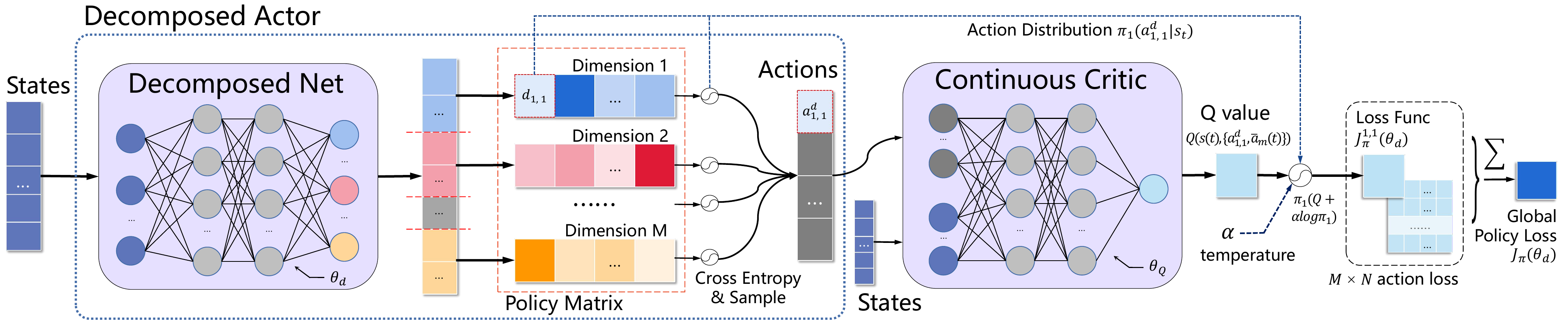}
	\caption{Computation flow of evaluating the loss function $J_{1,1}^{\pi}(\theta_d)$ in SDAC, with simultaneous evaluation of all $J_{m,n}^{\pi}(\theta_d)$ in a single step for improved computational efficiency.}
	\label{img2}
\end{figure}

First, we incorporate adaptive temperature adjustment by minimizing the loss function $J(\alpha)$, which constrains the entropy of our discrete policies to a target entropy $\widehat{\mathcal{H}}$:
\begin{equation}\label{alphaloss}
	J(\alpha)= \frac{1}{M} \sum_{m = 1}^{M}\mathbb{E}_{\pi_m}\!\big[-\alpha \log \pi_m(\bigcdot \ )|s(t);\theta_{d})-\alpha\widehat{{\mathcal{H}}} \big].
\end{equation}
However, achieving the target entropy $\widehat{\mathcal{H}}$ becomes challenging due to negative correlations between an appropriate $\widehat{\mathcal{H}}$ and the discrete actions $N$ for each dimension. To address this issue, we redefine $\mathcal{H}_m$ as follows (for more details, refer to Appendix \ref{appa})
\begin{equation}
	\label{practicalEntropy}
	\mathcal{H}_m(s(t);\theta_d)=\sum_{j=1}^N \pi_m(a^d_{m,n}|s(t);\theta_d) \ {\rm log} \ \big(\frac{2}{N}\pi_m(a^d_{m,n}|s(t);\theta_d)\big).
\end{equation}
For brevity, we have reformulated the policy loss functions in Equation \eqref{KLloss} as direct policy gradients, with their equivalence proven in Appendix \ref{appc2}. The updated $J^{\pi}(\theta_{d})$ now consists of sub-functions for each discrete action
\begin{align}
	\label{policyloss}
	J^{\pi}(\theta_{d}) & =\frac{1}{M}\sum_{m = 1}^{M}\sum_{n = 1}^{N}J^{\pi}_{m,n}(\theta_{d}) \ ,	\quad {\rm where}\\
	J^{\pi}_{m,n}(\theta_{d}) &=\pi_m(a^d_{m,n}|s(t);\theta_{d})\Big
	[\alpha\log \pi_m(a^d_{m,n}|s(t);\theta_{d})-Q(s(t), \{a^d_{m,n}, \overline{a}_m(t)\}; \theta_{Q}) \Big].\notag 	
\end{align}
Furthermore, we integrate the double $Q$-network \cite{van2016deep} and soft target update into our practical SDAC algorithm. The updates performed by SDAC are depicted in Figure \ref{img2}. For more detailed algorithm specifications, please refer to Appendix \ref{appa}. In Section \ref{evaluation}, we present experimental results that demonstrate the comparable or superior performance of SDAC to SAC in various environments.

\section{Soft decomposed-critic Q: Continuous values to discrete values}\label{SDCQ}

In this section, we introduce soft decomposed-critic Q (SDCQ), a novel algorithm that enhances the training efficiency of our previous SDPC architecture. To achieve this, we analyze the KL divergence policy loss function in Equation \eqref{KLloss} and replace it with the squared error between continuous $Q$-values \eqref{continuousQ} and discrete $Q$-values \eqref{discreteQ}. SDCQ utilizes a decomposed discrete $Q$-network and a continuous critic network, updating them through value iterations and supervised learning. Additionally, it generates soft Boltzmann exploration policies based on the discrete $Q$-values, with an adaptive temperature controlling the exploration rate. Through these modifications, SDCQ achieves remarkable performance compared to baseline algorithms.

\subsection{From policy gradients to supervised learning}
Previous studies \cite{nachum2017bridging, o2016combining, schulman2017equivalence} have bridged the gap between policy gradients and $Q$-learning in the context of soft RL. In this section, we provide further analysis of the KL divergence in Equation \eqref{KLloss}, which motivates the transition from policy gradients to supervised learning for $Q$-functions. We focus on an optimal discrete policy $\pi_m^*(a_m(t)|s(t);\theta_d^*)$ with an optimal parameter $\theta_d^*$, where the KL divergence is zero, i.e., $J^\pi_m(\theta_d^*)=0$. The optimal policy $\pi_m^*(a_m(t)|s(t);\theta_d^*)$ follows the distribution given by
\begin{equation}\label{optimalpolicy}	
	\pi_m(a^d_{m,n}|s(t);\theta_d)=\frac{\exp\big(\frac{1}{\alpha}Q(s(t),{a^d_{m,n},\overline{a}_m(t)};\theta_Q)\big)}{\sum_{j=1}^N \exp\big(\frac{1}{\alpha}Q(s(t),{a^d_{m,j},\overline{a}_m(t)};\theta_Q)\big)}.
\end{equation}
It is worth noting that Equation \eqref{optimalpolicy} resembles the cross-entropy function in Equation \eqref{policy}. We observe connections between the output $D(s(t);\theta_d^*)$ of the soft policy networks and the soft $Q$-values. Specifically, under the optimal policies, we have that
\begin{align}\label{DtoQ}	
	&\forall m\in[M];  \  j,k\in[N]:  \notag\\[-0.1cm]  
	 &d_{m,j}(s(t);\theta_d^*) - d_{m,k}(s(t);\theta_d^*) = \frac{1}{\alpha}\Big(Q_d(s(t), a^d_{m, j};\overline{\pi}^*_{m}) - Q_d(s(t), a^d_{m, k};\overline{\pi}^*_{m})\Big).
\end{align}
From $D(s(t);\theta_d^*)$, we obtain the corresponding advantage function of the discrete $Q$-values
\begin{equation}\label{advantage}
	A(s(t); a^d_{m, n})= \alpha\Big(d_{m,n}(s(t);\theta_d^*) - \sum_{j=1}^N \pi_m^*(a^d_{m,n}|s(t);\theta_d^*) d_{m,j}(s(t);\theta_d^*)\Big)
\end{equation}
The SDCQ algorithm is inspired by the relationship between decomposed policy networks and $Q$-values, which has led to a novel approach bridging the gap between continuous and discrete $Q$-values. This approach generates a maximum-entropy policy based on the decomposed $Q$-function. In SDCQ, the policy network used is the decomposed $Q$-network, which produces a matrix $D(s(t);\theta_d)$ consisting of decomposed $Q$-values
\begin{equation}\label{decomposedQnet}
	Q_d(s(t),a^d_{m,n};\theta_d)=d_{m,n}(s(t);\theta_d) .
\end{equation}  
To obtain the probability distribution of the decomposed actions, SDCQ employs a Boltzmann policy with an adaptive temperature parameter $\alpha$. This adaptive temperature allows precise control over the distribution of Boltzmann policies. The Boltzmann policy in SDCQ is defined as follows
\begin{equation}\label{boltzmann}
	\pi_m(a^d_{m,n}|s(t);\theta_d,\alpha)=\frac{\exp \big(\frac{1}{\alpha} Q_d(s(t),a^d_{m,n};\theta_d)\big)}{\sum_{j=1}^{N}\exp \big(\frac{1}{\alpha} Q_d(s(t),a^d_{m,j};\theta_d)\big)}.
\end{equation}
In SDCQ, the objective function $J^\pi_m(\theta_d)$ for the policy $\pi_m$ can be constructed using the KL divergence, similar to Equation \eqref{KLloss}. The unique minimum point of $J^{\pi}_m(\theta_{d})$ occurs when $Q_d(s(t),\bigcdot \ ;\theta_d)- Q(s(t),\{\bigcdot \,\overline{a}_m(t)\} ;\theta_Q)=\delta$  for any discrete action in dimension $m$, where $\delta\in \mathbb{R}$ is arbitrary. At this minimum point, $J^{\pi}_m(\theta_d)$ reduces to the variance $\sigma_{\pi_m}^2$ of $Q - Q_d$ among action dimension $m$
\begin{align}\label{supervisedequality}
	\lim\limits_{Q_d\rightarrow Q + \delta}J^{\pi}_m(\theta_d)=& \ \sigma_{\pi_m}^2\Big[Q_d\big(s(t),\bigcdot \ ;\theta_d\big)- Q\big(s(t),\{\bigcdot,\overline{a}_m(t)\};\theta_Q\big)\Big] \notag\\
	=& \ \mathbb{E}_{\pi_m}\big[(Q_d- Q)^2\big]-\big(\mathbb{E}_{\pi_m}(Q_d- Q)\big)^2.
\end{align}
This relationship is proven in Appendix \ref{appc3}. It is observed that the minimum of $\mathbb{E}_{\pi_m}\big[(Q_d- Q)^2\big]$ with $\delta=0$ is a sufficient condition for the minimum of $J^{\pi}_m(\theta_d)$, while there is no correlation between the minimum of $\big(\mathbb{E}_{\pi_m}(Q_d- Q)\big)^2$ and $J^{\pi}_m(\theta_d)$. Therefore, the KL divergence can be minimized indirectly through the following objective function
\begin{equation}\label{supervisedorigin}
	J^{d}_m(\theta_d) = \mathbb{E}_{\pi_m}\!\Big[\big(Q_d(s(t),\bigcdot \ ;\theta_d)- Q(s(t),\{\bigcdot,\overline{a}_m(t)\};\theta_Q)\big)^2\Big]
\end{equation}
which resembles the squared error loss, aiming to minimize the discrepancy between $Q_d$ and $Q$. By leveraging the relationship between $J^{\pi}_m(\theta_d)$ and $J^{d}_m(\theta_d)$, we can optimize the soft Boltzmann policy strictly through supervised learning, leading to our practical SDCQ algorithm.

\subsection{Practical algorithm}\label{practicalSDCQ}
To generalize learning across decomposed actions \cite{sewak2019deep}, we propose the practical SDCQ algorithm, which utilizes a decomposed $Q$-network to learn the discrete advantage function $A_d$
\begin{equation}\label{practicaladvantage}
	A_d(s(t);a^d_{m,n};\theta_d) = d_{m,n}(s(t);\theta_d) =  Q_d(s(t),a^d_{m,n};\theta_d) - \mathbb{E}_{\pi_m}\big[Q_d(s(t),\bigcdot \ ;\theta_d)\big].
\end{equation} 
As mentioned in Equation \eqref{supervisedequality}, the KL divergence loss function for optimizing soft policies is equivalent to the square loss function at its minimum point. To simplify our algorithm, we use the mean square error between the discrete and continuous advantage functions as our policy loss
\begin{align}\label{supervisedloss}
	J^{d}(\theta_{d})&=\frac{1}{M}\sum_{m = 1}^{M}\sum_{n = 1}^{N}J^{d}_{m,n}(\theta_{d}),\quad {\rm where}	 \\[-0.15cm]
	J^{d}_{m,n}(\theta_{d}) &= \Big(d_{m,n}(s(t);\theta_d) -\big(Q(s(t),\{a^d_{m,n},\overline{a}_m(t)\};\theta_Q) - \mathbb{E}_{\pi_m}[Q_d(s(t),\{\bigcdot \ , \overline{a}_m(t)\} ;\theta_Q)]\big)\Big)^2 \notag
\end{align} 
In the continuous soft $Q$-value iteration of SDCQ, presenting Equation  \eqref{valueits} strictly as our value loss function is problematic and unstable in the training process. Since the policy distribution of SDCQ is strictly controlled by $\alpha$, the rapidly changing adaptive temperature may disrupt the process of sampling the target action $\widetilde{a}_{t+1}$. In our practical SDCQ algorithm, we set a target temperature $\alpha'$ which undergoes soft target updates as $\alpha'\leftarrow (1-\tau)\alpha' + \tau\alpha, \tau \in (0,1)$ along with the target critic network. Additionally, a target decomposed Q network is not necessary. The target temperature $\alpha'$ is employed in the target policy $\pi'(\widetilde{a}(t+1)|s(t+1);\theta_d,\alpha')$  to generate target actions, along with its corresponding entropy $\mathcal{H}'_{t+1}=\mathcal{H}(s(t+1); \theta_d, \alpha')$  as defined in  Equation \eqref{entropy}.

Another ingredient that contributes to the strong performance of SDCQ is a multi-step temporal difference with importance sampling. Similar to Rainbow DQN \cite{hessel2018rainbow}, multi-step TD can enhance performance in a series of algorithms. We find it effective in SDCQ as well. We utilize a three-step value loss function, replacing Equation \eqref{valueits} in SDCQ, where $\widetilde{a}(t+3)$ is sampled from $\pi'$ as follows
\begin{align}\label{multistep}
	J^Q (\theta_Q)= \Big(Q(s(t), &a(t); \theta_{Q})
	- \Big(r_t +\gamma \big(\alpha \mathcal{H}_{t+1} + r_{t+1}\big) \\[-0.15cm]
	&+ \gamma^2\big(\mathcal{H}_{t+2} + r_{t+2}\big)
	+\gamma^3\big(\mathcal{H}_{t+3} + Q(s(t+3), \widetilde{a}(t+3); \theta_{Q}')\big)\Big)\Big)^2 \notag.
\end{align}
However, biases in the follow-up rewards after $r_t$ can introduce drawbacks in terms of robustness. To address this issue, we leverage the soft discrete policies and correct the biases through importance sampling. During the interaction process, we sample transition experiences $\{s(t),a(t), p_{\text{old}}(a(t)), r_t, s(t+1)\}$ and store them in our replay buffer $\mathcal{D}$, which includes the probability $p_{\text{old}} (a(t))$ of the actions. When performing value iterations, we weight our samples by the importance $I(t)$ of the follow-up steps
 \begin{subequations}
 \begin{align}\label{importance}
 	I(t) =  p_{old}\big(a(t)\big) \big/ \pi\big(a(t)|s(t);\theta_d;\alpha'\big).
 \end{align}
 The critic parameters $\theta_Q$ are updated by taking steps in the direction that minimize the following expectation over a mini-batch $\mathcal{D}^-$ sampled from $\mathcal{D}$
\begin{align}\label{paramiteration}
	\theta_Q\leftarrow\theta_Q - \lambda_Q  \  \mathbb{E}_{\mathcal{D}^-}\Big[I(t+1)I(t+2) \frac{\partial J^Q(\theta_Q)} { \partial \theta_Q}\Big].
\end{align}
\end{subequations}
Here, $\lambda_Q$ represents the step size. It is worth noting that there are considerations regarding the use of importance sampling in offline RL, as $I(t)$ may become invalid due to the discrepancy between old policies and the current policy. To address this issue, we employ normalized importance sampling in SDCQ. Specifically, we compute the standard deviation $\sigma_{\mathcal{D}^-}$ of $\log I(t)$ among $\mathcal{D}^-$, and set a hyperparameter $\sigma$ to $2$ in general. The importance weights are then updated as follows
\begin{equation}\label{importancenorm}
	I(t)\leftarrow \exp\Big(\sigma \cdot \rm{clip}\big(\frac{\log \emph{I}(t) - \mathbb{E}_{\mathcal{D}^-}[\log \emph{I}(\bigcdot)]}{\sigma_{\mathcal{D}^-}[\log \emph{I}(\bigcdot)]},  \ -1, \ 1 \big)\Big).
\end{equation}
In conclusion, to learn the parameters of the critic in SDCQ, we employ multi-step soft value iterations with normalized importance sampling, while the target temperature is set to improve stability. The parameters of the decomposed $Q$-network, which can be viewed as a decomposed advantage network, are optimized through supervised learning. With these ingredients, SDCQ demonstrates outstanding performance in our benchmarks and achieves significantly accelerated training speed compared to the baseline algorithms. We present the additional details of SDCQ in Appendix \ref{appb}.

\section{Experimental evaluation}\label{evaluation}

In this section, we present our experimental evaluation to assess the effectiveness and efficiency of our proposed discrete RL methods on challenging continuous control tasks from the OpenAI Gym \cite{brockman2016openai} benchmark suite. These tasks are based on MuJoCo \cite{todorov2012mujoco} and Box2d \cite{parberry2017introduction} environments and include Hopper-v2, Walker2d-v2, Ant-v2, Humanoid-v2, InvertedDoublePendulum-v2, and BipedalWalker-v3. Appendix \ref{appd} provides detailed information about the baseline environments.

To evaluate the performance of our algorithms, we compare them to state-of-the-art off-policy actor-critic algorithms, namely SAC \cite{haarnoja2018soft2} and TD3 \cite{fujimoto2018addressing}. We also include the offline discrete RL algorithm BDQ \cite{tavakoli2018action}, which discretizes continuous actions into separate discrete branches to solve continuous control tasks. We use the author-provided implementations of the baseline actor-critic algorithms with their original hyperparameters, except for TD3, where we set the Gaussian exploration noise $\sigma=0.05$ in Humanoid-v2 to prevent training failures. For BDQ, we use our own implementation. The hyperparameters for the baseline algorithms are set to their default values. Each action dimension is discretized into $N=20$ actions, with continuous actions clipped to the range $[-1, 1]$. We set the target entropy as $\widehat{\mathcal{H}}=-1$ for SDAC and $\widehat{\mathcal{H}}=0$ for SDCQ to balance exploration and exploitation. Additional network details and algorithm hyperparameters are provided in Appendices \ref{appa} and \ref{appb}.

All algorithms are implemented in PyTorch \cite{paszke2019pytorch} and evaluated on a personal computer with an NVIDIA RTX3080TI. We conduct multiple training runs for each task, using the Adam optimizer \cite{kingma2014adam}, and report the experimental results averaged over 5 random seeds. During the training process, we evaluate the algorithm's performance over five episodes by turning off the exploration strategy every few steps ($1,000$ for InvertedDoublePendulum-v2 and BipedalWalker-v3, $5,000$ for the others).

\begin{figure}[t]
	\centering
	\includegraphics[width=5.5in]{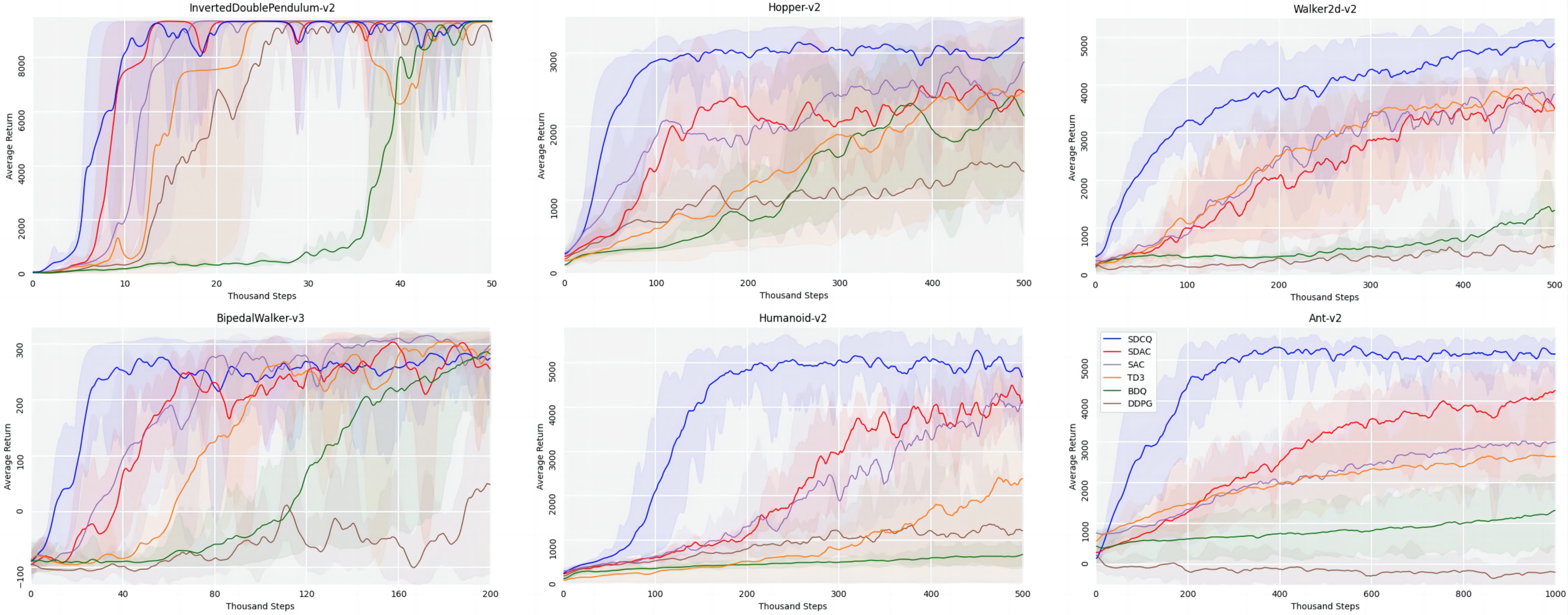}
\caption{Training curves of our proposed SDAC, SDCQ and the baseline algorithms on benchmark tasks, averaged over $5$ random seeds.}
	\label{img7}
\end{figure}

Figure \ref{img7} shows smoothed total average reward curves to provide a clear visualization of the experimental results. Surprisingly, the proposed SDAC with discrete soft policies demonstrates similar or even competitive training performance compared to the state-of-the-art continuous RL methods across all benchmark environments. In contrast, existing discrete RL methods like BDQ can only compete with the DDPG method \cite{lillicrap2015continuous}, but are inferior to TD3 and SAC.

Building on the advances of SDAC, SDCQ achieves even better results, converging with fewer training steps (less than $1/2$ compared to SAC). Moreover, SDCQ can owns the highest final performances among all six algorithms. From these perspectives, our discrete RL methods achieve significantly improved performance over actor-critic methods in these continuous control tasks. Additionally, we conduct the ablation studies for our algorithms in Appendix \ref{appe} due to space limitations.

\section{Conclusions}

In this work, we have addressed the challenge of applying RL to continuous tasks by introducing soft decomposed discrete policies. We proposed a decomposed policy-critic architecture that tackles the dimensional explosion problem through independent action-dimensional discretization. Furthermore, we improved the training efficiency by offering a continuous critic network and bridging the gap between discrete Q and continuous Q functions. Based on the SDPC architecture, we developed two algorithms: SDAC with policy gradients and SDCQ, which learns optimal soft policies through supervised learning.
Our extensive experiments on benchmark tasks have demonstrated the effectiveness and efficiency of our discrete RL algorithms compared to competing continuous RL approaches such as SAC, TD3, DDPG, as well as the discrete RL method BDQ. Our algorithms exhibit higher training efficiency, along with outstanding final performances.

While our proposed methods have shown promising results, there are several directions for further improvement. Firstly, the generation of discrete action $Q$-values through linear interpolation instead of relying on the critic network can reduce computational complexity. Additionally, exploring discrete-continuous hybrid algorithms that train the continuous actor using the critic in SDPC could be an effective approach to improve decision speed during interactions with the environment.

\clearpage

\bibliographystyle{IEEEtranS}
\bibliography{references1.bib}

\clearpage

\appendix
\renewcommand{\thesection}{Appendix A}

\section{Implementation Details of SDAC}\label{appa}

\subsection{Entropy Normalization in SDAC}
In SDAC, The temperature adjustment in Equation \eqref{alphaloss} aims to align the expectations of $\mathcal{H}_m$ with the target entropy $\hat{\mathcal{H}}$ for each action dimension, which means that our target entropy $\hat{\mathcal{H}}$ do not needs to take adjustments corresponding to the action dimensions $dim(\mathcal{A})$. However, we must concern about the the impact of action discretization accuracy $N$ on the target entropy $\hat{\mathcal{H}}$. 

\begin{figure}[H]
	\centering
	\includegraphics[width=5in]{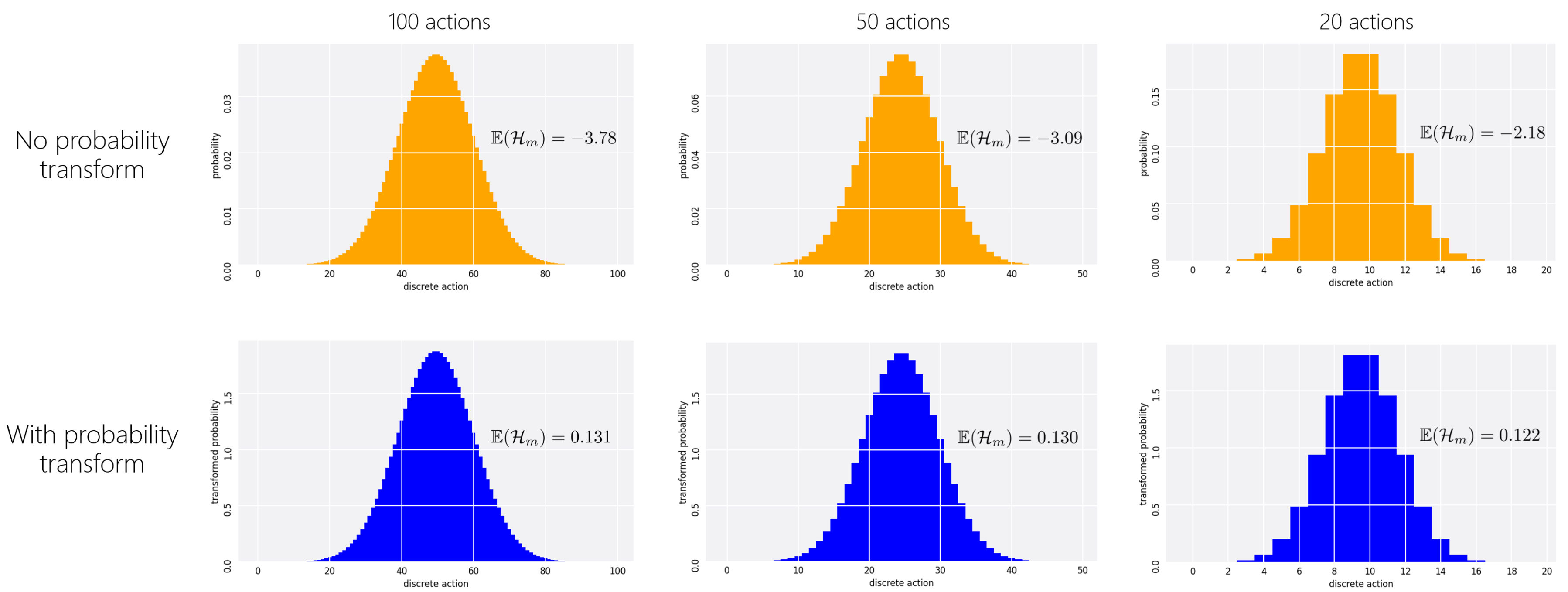}
	\caption{Effect of entropy normalization on the sensitivity of $\mathbb{E}(\mathcal{H}_m)$ to discrete actions $N$.}
	\label{entropytransfer}
\end{figure}

Consider the original entropy for the discrete policies on each action dimension:
\begin{equation}
	\mathcal{H}_m=\sum_{n=1}^{N}-\pi_m(a^d_{m,n}|s(t);\theta_{d})\log  \pi_m(a^d_{m,n}|s(t);\theta_{d})
\end{equation}
When comparing two discretized policies with similar distributions but different action fractions $N$, their respective $\mathcal{H}_m$ values may exhibit significant differences, making it challenging to achieve the desired target entropy $\hat{\mathcal{H}}$. To illustrate this, consider a continuous stochastic policy parameterized by a Gaussian distribution with $\mu=0$ and $\sigma=0.3$ defined on the interval $[-1, 1]$. When we transfer this continuous policy to discrete settings, we obtain $\mathbb\mathcal{H}_m=-3.78$ for $N=100$, $\mathcal{H}_m=-3.09$ for $N=50$, and $\mathcal{H}_m=-2.18$ for $N=20$, as shown in Figure \ref{entropytransfer}.

To address the challenge of adjusting the target entropy $\hat{\mathcal{H}}$ in each action dimension, we propose an entropy normalization method in SDAC and SDCQ, which transforms discrete policy distributions into probability densities on the continuous action space. In typical continuous control tasks where actions are clipped to the range $[-1, 1]$, discretizing an action dimension uniformly into $N$ discrete actions results in each discrete action representing a fraction of length $2/N$ on the continuous action space. By applying probability transfer, the probability $\pi_m(a^d_{m,n}|s(t);\theta_{d})$ is transformed into a probability density $2\pi_m(a^d_{m,n}|s(t);\theta_{d})/N$. Consequently, the entropy $\mathcal{H}_m$ on each action dimension becomes
\begin{equation}
	\mathcal{H}_m=\sum_{n=1}^{N}-\pi_m(a^d_{m,n}|s(t);\theta_{d})\log \big[\frac{2}{N} \pi_m(a^d_{m,n}|s(t);\theta_{d})\big]
\end{equation}

Revisiting the previous example, with probability transfer, we obtain $\mathcal{H}_m = 0.131$ for $N=100$, $\mathcal{H}_m = 0.130$ for $N=50$, and $\mathcal{H}_m = 0.122$ for $N=20$, demonstrating stability with respect to $N$. Thus, the probability transfer method presented in SD2PC effectively reduces the complexity of adjusting $\hat{\mathcal{H}}$, resulting in our algorithm being less sensitive to hyperparameters.
\clearpage

\subsection{Pseudo code of Algorithm SDAC}\label{appa1}

To address the over-estimation problem of the critic network and improve the performance of our algorithm, we incorporate the double critic network from TD3\cite{fujimoto2018addressing} into SDAC. In our practical SDAC algorithm, we parameterize $\theta_{Q1}$ and $\theta_{Q2}$ for a pair of critic networks, as well as $\theta'_{Q1}$ and $\theta'_{Q2}$ for their respective target networks. These two networks have independent training processes but share a target Q value function computed using minimum selection. Thus, the Q value loss function in Eq.\eqref{valueits} is replaced by:
\begin{align}
	\label{doublevalueits}
	J^Q (\theta_{Q_i}) = \Big[Q(s(t), a(t); \theta_{Q}) - \Big(r_t &+\gamma \big(\alpha \mathcal{H}(s(t+1);\theta_d)\notag\\[-0.25cm]
	&+\min_{j=\{1,2\}}Q(s(t+1), \widetilde{a}(t+1); \theta_{Q_j}')\big)\Big) \Big]^2.
\end{align}
Here, $i\in\{1,2\}$ and $\widetilde{a}(t+1)$ is sampled from the policy $\pi(a(t)|s(t);\theta_d)$ Eq.\eqref{policy}. When evaluating continuous Q values for the KL divergence policy loss function in Eq.\eqref{policyloss}, SDAC use the minimum value between these two critics. The policy loss based on the double critic networks is defined as
\begin{align}
	\label{doublepolicyloss}
	J^{\pi}(\theta_{d}) & =\frac{1}{M}\sum_{m = 1}^{M}\sum_{n = 1}^{N}J^{\pi}_{m,n}(\theta_{d}) \ ,	\quad {\rm where}\\[-0.1cm]
	J^{\pi}_{m,n}(\theta_{d}) &=\pi_m(a^d_{m,n}|s(t);\theta_{d})\Big
	[\alpha\log \pi_m(a^d_{m,n}|s(t);\theta_{d})-\min_{j=\{1,2\}}Q(s(t), \{a^d_{m,n}, \overline{a}_m(t)\}; \theta_{Q_j}) \Big].\notag 	
\end{align}
Furthermore, SDAC use soft targt updates to iterate the parameters of the target critic networks with a tiny  $\tau$ in each training step:
\begin{align}
	\theta_{Q1}'\leftarrow \tau \theta_{Q1} +  (1 - \tau) \theta_{Q1}\\
	\theta_{Q2}'\leftarrow \tau \theta_{Q2} +  (1 - \tau) \theta_{Q2}''\notag
\end{align}
By replacing Eq. \eqref{valueits} and \eqref{policyloss} with the value and policy loss functions in Eq. \eqref{doublevalueits} and \eqref{doublepolicyloss}, we present the pseudo code for our SDAC algorithm with double critic networks. Here, $\lambda_d$, $\lambda_Q$, and $\lambda_\alpha$ represent the learning rates for the decomposed policy network, critic networks, and adaptive temperature $\alpha$ respectively. 
\begin{algorithm}[H]
	\caption{Soft Decomposed Actor-Critic (SDAC)}
	\setstretch{1.05}
	\begin{algorithmic}[H] 
		\Require network parameters $\theta_{d}$, $\theta_{Q1}$, $\theta_{Q2}$, target network parameters  $\theta_{Q1}'$, $\theta_{Q2}'$
		\Require replay buffer $\mathcal{D}\leftarrow \{ \}$, temperature $\alpha$, hyperparameters $\lambda_d, \ \lambda_Q, \ \lambda_\alpha,\ \tau, \ \gamma$
		\Require target entropy $\hat{\mathcal{H}}$
		\For {$t=0$ \textbf{to} $T$}
		\State Sample an action from policy distribution: $a(t) \sim \pi(a(t)|s(t);\theta_{d})$
		\State Takes action $a(t)$, observes reward $r_t$ and next state $s(t+1)$ 
		\State Store transition experience $\{s(t), a(t), r_t, s(t+1)\}$ in $\mathcal{D}$
		\If{Training}
		\State Sample a minibatch $\mathcal{D}^-$ from $\mathcal{D}$
		\State $\theta_{Qi} \leftarrow \theta_{Qi} - \lambda_Q\nabla_{\theta_{Qi}} J^{Q} (\theta_{Qi})$ \   with Eq.\eqref{doublevalueits}
		\State $\theta_{d} \leftarrow \theta_{d}-\lambda_d\nabla_{\theta_{d}} J^\pi (\theta_{d})$ with Eq.\eqref{doublepolicyloss}
		\State $\alpha \leftarrow \alpha-\lambda_\alpha\nabla_{\alpha} J(\alpha)$ 
		\State  $\theta_{Q1}'\leftarrow \tau \theta_{Q1} +  (1 - \tau) \theta_{Q1}'$
		\State  $\theta_{Q2}'\leftarrow \tau \theta_{Q2} +  (1 - \tau) \theta_{Q2}'$
		\EndIf
		\EndFor
	\end{algorithmic}
\end{algorithm}

\clearpage

\subsection{Neural Network Structure}\label{appa2}

As discussed in Section \ref{SDAC}, our SDPC structure contains two types of networks, the decomposed policy networks and the continuous critic networks. These two types of networks are both based on MLPs, which the policy network contains two hidden layers with 256 neurons, and an output layer with $M\times N$ neurons which can reconstruct to the policy matrix. Similarly the continuous critic network have two hidden layers with 256 neurons, but with a single neuron which outputs the continuous Q value $Q(s(t), a(t);\theta_Q)$ corresponding to the input states and actions.

\begin{figure}[H]
	\centering
	\includegraphics[width=5.5in]{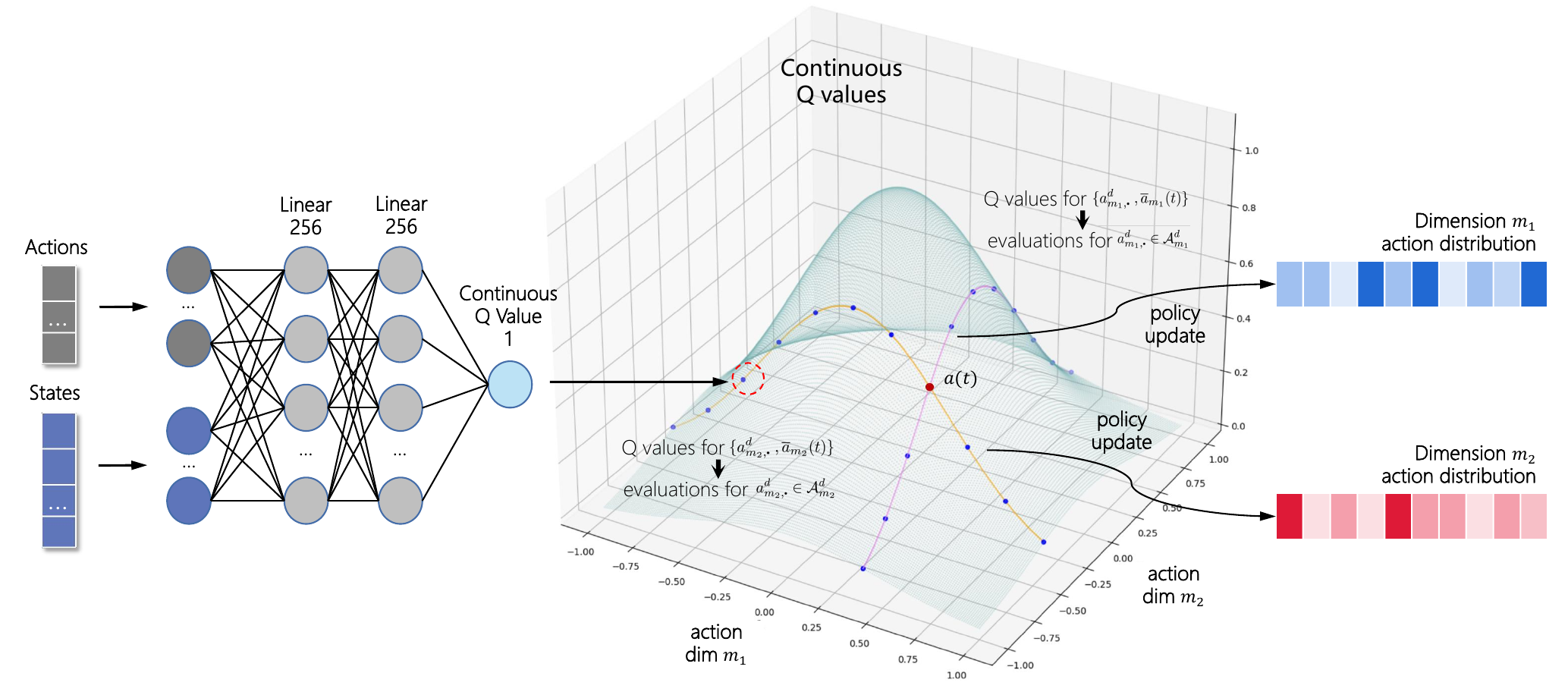}
\caption{Critic network and policy update details in SDAC. Each policy update for a single action dimension requires $N$ forward propagations of the critic network. In the global policy update, $M \times N$ forward propagations are performed, ensuring computational efficiency.}
	\label{networkstructure}
\end{figure}

Attention that the critic network represents a continuous state-action value function, which means we can only obtain the value of a single discrete action point at a time, as depicted by the dots on the critic value function in Figure \ref{networkstructure}. Therefore, we need to forward the critic network $N$ times for each action dimension's policy update, and $M \times N$ times for the global policy loss function. These $M \times N$ Q values can be calculated simultaneously in a minibatch, resulting in reasonable computational complexity for the policy update in SDAC and SDCQ.

\subsection{Hyperparameters for SDAC}

\begin{table}[H]
	\caption{Hyperparameters of SDAC}
	\label{t1}
	\begin{center}
		\begin{tabular}{p{60pt}l|ll} 
			
			\toprule 
			Hyperparameter & & Definition & Value  \\
			
			\midrule 
			$N$   & & discrete actions per dimension    &  20\notag \\[0.05cm] 
			$\hat{\mathcal{H}}$   & & target entropy     &  -1.5\notag \\[0.05cm]    
			$\tau$ & & stepsize of soft target update  & $5\times 10^{-3}$ \notag \\[0.05cm]
			$\gamma$ & & reward discounting rate  & $0.99$ \notag \\[0.05cm]
			$\mathcal{D}$       & & buffer size           & 256  \notag \\[0.05cm]
			$\mathcal{D}^-$       & & batch size           & 256  \notag \\[0.05cm]
			$\max(\log \ \alpha)$  & & temperature upper bound      &  2     \notag \\[0.05cm]
			$\min(\log \ \alpha)$  & & temperature lower bound        &  -10   \notag \\[0.05cm]
			$\lambda_\alpha$       & & temperature learning rate           & $3\times 10^{-4}$ \notag \\[0.05cm]
			\multirow{2}[2]{*}{$\lambda_d$} & &  \multirow{2}[2]{*}{policy learning rate} & $3\times 10^{-4}$ (Humanoid-v2, Ant-v2) \\
			& &                  &  $10^{-3}$ (other environments)            \notag \\[0.05cm]
			\multirow{2}[2]{*}{$\lambda_Q$} & &  \multirow{2}[2]{*}{critic learning rate} & $3\times 10^{-4}$ (Humanoid-v2, Ant-v2) \\
			& &                  &  $10^{-3}$ (other environments)            \\
			\bottomrule 
		\end{tabular}
	\end{center}
\end{table}

\clearpage
\renewcommand{\thesection}{Appendix B}
\section{Implementation Details for SDCQ}\label{appb}
\subsection{Pseudocode of Algorithm SDCQ}
Building on the advances of SDAC, SDCQ inherits some techniques which presents in Appendix.\ref{appa}, including entropy normalization, soft target update and the network structures. The 3-step TD critic loss function for SDCQ has transferred to the follwing object:
\begin{align}\label{doublemultistep}
	J^Q (\theta_{Q_i})&= \Big((s(t), a(t); \theta_{Q_i})
	- \Big(r_t +\gamma \big(\alpha \mathcal{H}_{t+1} + r_{t+1}\big) \\[-0.15cm]
	&+ \gamma^2\big(\mathcal{H}_{t+2} + r_{t+2}\big)
	+\gamma^3\big(\mathcal{H}_{t+3} + \min_{j=\{1,2\}}Q(s(t+3), \widetilde{a}(t+3); \theta_{Q_j}')\big)\Big)\Big)^2 \notag.
\end{align}
And the supervised policy loss of SDCQ under the double critics:
\begin{align}\label{doublesupervisedloss}
	J^{d}(\theta_{d})&=\frac{1}{M}\sum_{m = 1}^{M}\sum_{n = 1}^{N}J^{d}_{m,n}(\theta_{d}),\quad {\rm where}	 \\[-0.15cm]
	J^{d}_{m,n}(\theta_{d}) &= \Big(d_{m,n}(s(t);\theta_d) -\min_{j=\{1,2\}}\big(Q(s(t),\{a^d_{m,n},\overline{a}_m(t)\};\theta_{Q_j})  \notag\\
	&\quad \quad \quad \quad \quad \quad \quad \ \  \  -\mathbb{E}_{\pi_m}[\min_{j=\{1,2\}}Q(s(t),\{\bigcdot \ , \overline{a}_m(t)\} ;\theta_{Q_j})]\big)\Big)^2 \notag
\end{align}  
Furthermore, in SDCQ, besides the critic networks, we also present the soft target update on the target adaptive temperature $\alpha'$ with $\alpha'\leftarrow \tau \alpha +  (1 - \tau) \alpha'$.By replacing Eq.\eqref{supervisedloss} and \eqref{multistep} with the double supervised loss functions in Eq.\eqref{doublesupervisedloss} and \eqref{doublemultistep} , we present the pseudo code for our practical SDCQ algorithm as follows. The hyperparameters have the same meanings with SDAC.  
\begin{algorithm}[H]
	\caption{Soft Decomposed-Critic Q (SDCQ)}
	\setstretch{1.1}
	\begin{algorithmic}[H] 
		\Require network parameters $\theta_{d}$, $\theta_{Q1}$, $\theta_{Q2}$, target network parameters  $\theta_{Q1}'$, $\theta_{Q2}'$
		\Require replay buffer $\mathcal{D}\leftarrow \{ \}$, temperature $\alpha$,$\alpha'$, hyperparameters $\lambda_d, \ \lambda_Q, \ \lambda_\alpha,\ \tau, \ \gamma$
		\Require target entropy $\hat{\mathcal{H}}$, importance sampling deviation $\sigma$
		\For {$t=0$ \textbf{to} $T$}
		\State Generate Boltzmann policies $\pi_m(a^d_{m,n}|s(t);\theta_d,\alpha)$ from $D(s(t);\theta_d)$
		\State Sample an action from Boltzmann policy: $a(t) \sim \pi(a(t)|s(t);\theta_{d})$
		\State Store the sample probability $p_{old}(a(t))$ of $a(t)$
		\State Takes action $a(t)$, observes reward $r_t$ and next state $s(t+1)$ 
		\State Store transition experience $\{s(t), a(t),p_{old}(a(t)), r_t, s(t+1)\}$ in $\mathcal{D}$
		\If{Training}
		\State Sample a minibatch $\mathcal{D}^-$ from $\mathcal{D}$ with each experience contains 3 steps
		\State generate importance $\emph{I}(t+1),\emph{I}(t+2) $ with Eq.\eqref{importance},\eqref{importancenorm}
		\State $\theta_{Q_i}\leftarrow\theta_{Q_i} - \lambda_Q  \  \mathbb{E}_{\mathcal{D}^-}\Big[I(t+1)I(t+2) \frac{\partial J^Q(\theta_{Q_i})} { \partial \theta_{Q_i}}\Big]$ \  \textbf{For} $i\in\{1,2\} $ with Eq.\eqref{doublemultistep}
		\State $\theta_{d} \leftarrow \theta_{d}-\lambda_d\nabla_{\theta_{d}} J^d (\theta_{d})$ with Eq.\eqref{doublesupervisedloss}
		\State $\alpha \leftarrow \alpha-\lambda_\alpha\nabla_{\alpha} J(\alpha)$
		\State $\alpha'\leftarrow \tau \alpha +  (1 - \tau) \alpha'$
		\State  $\theta_{Q1}'\leftarrow \tau \theta_{Q1} +  (1 - \tau) \theta_{Q1}'$; $\theta_{Q2}'\leftarrow \tau \theta_{Q2} +  (1 - \tau) \theta_{Q2}'$
		\EndIf
		\EndFor
	\end{algorithmic}
\end{algorithm}

\clearpage
\subsection{Hyperparameters of  SDCQ}
Our advanced algorithm SDCQ owns similar hypermarameters with SDAC, including learning rates, discounting rate, buffer size and so on. It's worth mentioned that the ideal target entropy of our SDCQ algorithm is $\hat{\mathcal{H}}=0$ instead of $\hat{\mathcal{H}}=-1.5$ in SDAC, which leads to stronger exploration. In our ablation studies, we found that a lower $\hat{\mathcal{H}}$ may lead to drawbacks on the normalized importance sampling in SDCQ, so we recommanded to present $\epsilon-{\rm greedy}$ in SDCQ if it's necessary to reduce the exploration rate. 

Here we also presents the hyperparameters for the normalized importance sampling, including the upper bound, lower bound, and the log importance deviation $\sigma$. 

\begin{table}[H]
	\caption{Hyperparameters of SDCQ}
	\label{t2}
	\begin{center}
		\begin{tabular}{p{60pt}l|ll} 
			
			\toprule 
			Hyperparameter & & Definition & Value  \\
			
			\midrule 
			$N$   & & discrete actions per dimension    &  20\notag \\[0.15cm] 
			$\hat{\mathcal{H}}$   & & target entropy     &  -1.5\notag \\[0.15cm]    
			$\tau$ & & stepsize of soft target update  & $5\times 10^{-3}$ \notag \\[0.15cm]
			$\gamma$ & & reward discounting rate  & $0.99$ \notag \\[0.15cm]
			$\mathcal{D}$       & & buffer size           & 256  \notag \\[0.15cm]
			$\mathcal{D}^-$       & & batch size           & 256  \notag \\[0.15cm]
			$\max(\log \ \alpha)$  & & temperature upper bound      &  2     \notag \\[0.15cm]
			$\min(\log \ \alpha)$  & & temperature lower bound        &  -10   \notag \\[0.15cm]
			$\lambda_\alpha$       & & temperature learning rate           & $3\times 10^{-4}$ \notag \\[0.15cm]
			\multirow{2}[2]{*}{$\lambda_d$} & &  \multirow{2}[2]{*}{policy learning rate} & $3\times 10^{-4}$ (Humanoid-v2, Ant-v2) \\
			& &                  &  $10^{-3}$ (other environments)            \notag \\[0.15cm]
			\multirow{2}[2]{*}{$\lambda_Q$} & &  \multirow{2}[2]{*}{critic learning rate} & $3\times 10^{-4}$ (Humanoid-v2, Ant-v2) \\
			& &                  &  $10^{-3}$ (other environments)            \notag \\[0.15cm]
			$\sigma$ & & log importance deviation  & 2 \notag \\[0.15cm]
			$\max(\emph{I}(t)/ \sigma)$       & & importance upper bound           & 1  \notag \\[0.15cm]
			$\min(\emph{I}(t)/ \sigma)$       & & importance lower bound          & -1  \notag \\[0.15cm]
			\bottomrule 
		\end{tabular}
	\end{center}
\end{table}

\clearpage

\renewcommand{\thesection}{Appendix C}
\section{Proofs}\label{appc}

\subsection{Bridge between Discrete and Continuous Q-Value Functions}\label{appc1}
In section.\ref{MDPbridge}, we proposed a decomposed MDP model ${\mathcal{S}, \mathcal{A}^d_m, p_m, r_m}$ for the discrete policies on each action dimension, We have demonstrated that the decomposed MDP model holds the following relationship with the original continuous MDP model ${\mathcal{S}, \mathcal{A}, p_m, r_m}$:
\begin{subequations}
	\begin{align}
		&p_m\big(s(t+1)|s(t), a_m(t);\overline{\pi}_m\big) =  \ p\big( s(t+1)|s(t), \{a_m(t), \overline{a}_m(t)\} \big) \notag\\
		&r_m\big(s(t), a_m(t);\overline{\pi}_m\big) = \ r\big(s(t), \{a_m(t), \overline{a}_m(t)\} \big) + \alpha\overline{\mathcal{H}}_m\big(s(t), \overline{\pi}_m\big)\notag\\[-0.1cm]
		&{\rm where} \quad \overline{a}_m(t)\sim\overline{\pi}_m\big(\overline{a}_m(t) | s(t)\big),\quad \overline{\mathcal{H}}_m\big(s(t), \overline{\pi}_m\big) = \sum_{i=1, i\neq m}^{M}\mathcal{H}_i\notag.
	\end{align}
\end{subequations}
Which ``exclusive policy'' $\overline{\pi}_m=\{\pi_i| {i\in [M], i\neq m} \}$ as the set of discrete policies in each action dimension except for $m$. Based on these objects, the discrete soft state-action value function based on the decomposed MDP model ${\mathcal{S}, \mathcal{A}^d_m, p_m, r_m}$ can be transferred to continuous Q values through the following bridge:
\begin{align}
	\mathbb{E}_{a_m(t)\sim\pi_m}\big[Q_d(s(t), a_m(t);\overline{\pi}_m)\big] = \mathbb{E}_{a(t)\sim\pi}\big[Q(s(t), a(t))\big].\notag 
\end{align}

\begin{proof}
The decomposed Q value function is defined as the following object:
\begin{align}
	Q_d(s(t), a_m(t); \overline{\pi}_m) &= r_m(s(t),a_m(t);\overline{\pi}_m)    \\[-0.05cm]
	&\quad+\mathbb{E}_{p_m, \pi_m}\Big[\sum_{k=1}^{\infty}\gamma^k \big(r_m(s(t+k), a_m(t+k)) + \alpha\mathcal{H}_m(s(t+k))\big)\Big]\notag\\
	 &= \mathbb{E}_{\overline{\pi}_m}\big[r\big(s(t), \{a_m(t), \overline{a}_m(t)\} \big) + \alpha\overline{\mathcal{H}}_m\big(s(t)\big)\big] \notag\\
	&\quad+\mathbb{E}_{ p_m,  \pi_m}\Big[\sum_{k=1}^{\infty}\gamma^k \big(\mathbb{E}_{\overline{\pi}_m}\big(r(s(t+k), \{a_m(t+k), \overline{a}_m(t+k)\} ) \notag\\
	&\quad+\alpha\overline{\mathcal{H}}_m(s(t+k))\big)
	+ \alpha\mathcal{H}_m(s(t+k))\big)\Big]\notag\\
	&= \mathbb{E}_{\overline{\pi}_m}\big[r\big(s(t), \{a_m(t), \overline{a}_m(t)\} \big) + \alpha\overline{\mathcal{H}}_m\big(s(t)\big)\big] \notag\\
	&\quad+\mathbb{E}_{ p_m,  \pi_m}\Big[\sum_{k=1}^{\infty}\gamma^k \big(\mathbb{E}_{\overline{\pi}_m}\big(r(s(t+k), \{a_m(t+k), \overline{a}_m(t+k)\} )\big) \notag\\
	&\quad+\alpha\mathbb{E}_{\overline{\pi}_m}\big(\overline{\mathcal{H}}_m(s(t+k))
	+ \mathcal{H}_m(s(t+k))\big)\big)\Big]\notag\\
	&= \mathbb{E}_{\overline{\pi}_m}\big[r\big(s(t), \{a_m(t), \overline{a}_m(t)\} \big) + \alpha\overline{\mathcal{H}}_m\big(s(t)\big)\big] \notag\\
	&\quad+\mathbb{E}_{ p_m, \pi_m, \overline{\pi}_m}\Big[\!\sum_{k=1}^{\infty}\!\gamma^k \big(r(s(t\!+\!k), \{a_m(t\!+\!k), \overline{a}_m(t\!+\!k)\!\} )\!+\!\alpha\mathcal{H}(s(t\!+\!k))\big)\!\Big]\notag
\end{align}
\clearpage
Consider $\pi=\{\pi_m, \overline{\pi}_m\}$, $a(t)=\{a_m(t), \overline{a}_m(t)\}$, and $p_m$ is a function of $\overline{\pi}_m$; we can do further simplifications on $Q_d(s(t), a_m(t); \overline{\pi}_m)$:
\begin{align}\label{MDPproof}
	Q_d(s(t), a_m(t); \overline{\pi}_m)	&= \mathbb{E}_{\overline{\pi}_m}\big[r\big(s(t), a(t) \big)\big] + \alpha\mathbb{E}_{\overline{\pi}_m}\big[\overline{\mathcal{H}}_m\big(s(t)\big)\big] \notag\\
	&\quad+\mathbb{E}_{p, \pi}\Big[\sum_{k=1}^{\infty}\gamma^k \big(r(s(t+k), a(t+k))+\alpha\mathcal{H}(s(t+k))\big)\Big]\notag\\
	&=\mathbb{E}_{\overline{\pi}_m}\big[Q(s(t), a(t))\big]+\alpha\mathbb{E}_{\overline{\pi}_m}\big[\overline{\mathcal{H}}_m\big(s(t)\big)\big];\notag\\[0.4cm]
	\mathbb{E}_{\pi_m}\big[Q_d(s(t), a_m(t); \overline{\pi}_m)\big]&=\mathbb{E}_{\pi_m}\big[\mathbb{E}_{\overline{\pi}_m}\big[Q(s(t), a(t))\big]\big]+\alpha\mathbb{E}_{\pi_m}\big[\mathbb{E}_{\overline{\pi}_m}\big[\overline{\mathcal{H}}_m\big(s(t)\big)\big]\big]\notag\\[0.15cm]
	&=\mathbb{E}_{\pi}\big[Q(s(t), a(t))\big]+\alpha\mathbb{E}_{\overline{\pi}_m}\big[\overline{\mathcal{H}}_m\big(s(t)\big)\big]
\end{align}

From Eq.\eqref{MDPproof} we observed that $\alpha\mathbb{E}_{\overline{\pi}_m}[\overline{\mathcal{H}}_m(s(t))]$ is independent to the single-dimension discrete policy $\pi_m$. To reduce the computational complexity of our methods, we can ignore the exclusive entropy object in Eq.\eqref{MDPproof}, which leads to the bridge presented in Eq.\eqref{Q-bridge}.
\end{proof}
\clearpage

\subsection{KL divergence to policy gradients loss function}\label{appc2}
In section.\ref{MDPbridge}, we proposed the straight policy gradients loss function \eqref{policyloss} to optimize the decomposed soft policies in SDAC. In this section we proved which the policy loss is equal to the KL divergence object in Eq.\eqref{KLloss}.
\begin{proof}
\begin{align}
	\label{KLprove}
J^{\pi}_m(\theta_{d}) 
&={\rm D_{KL}} \bigg(\pi_m(\bigcdot 
\ |s(t);\theta_{d})\Bigg|\Bigg|\frac{\exp\big(\frac{1}{\alpha}Q(s(t),\{\bigcdot \ ,\overline{a}_m(t)\};
	\theta_Q)\big)}{\sum_{j=1}^{N}\exp\big(\frac{1}{\alpha}Q(s(t),\{a^d_{m,j} ,\overline{a}_m(t)\};\theta_Q)\big)} \bigg)\notag\\
&=\sum_{n=1}^N\pi_m(a^d_{m,n}|s(t);\theta_d)\Big(\log\pi_m(a^d_{m,n}|s(t);\theta_d) \notag\\ 
&\quad -\log\frac{\exp\big(\frac{1}{\alpha}Q(s(t),\{a^d_{m,n} ,\overline{a}_m(t)\};\theta_Q)\big)}{\sum_{j=1}^{N}\exp\big(\frac{1}{\alpha}Q(s(t),\{a^d_{m,j} ,\overline{a}_m(t)\};\theta_Q)\big)}\Big)\notag\\	
&=\sum_{n=1}^N\pi_m(a^d_{m,n}|s(t);\theta_d)\Big(\log\pi_m(a^d_{m,n}|s(t);\theta_d)-\frac{1}{\alpha}Q\big(s(t),\{a^d_{m,n} ,\overline{a}_m(t)\};\theta_Q\big) \notag\\ 
&\quad +\log\big[\sum_{j=1}^{N}\exp\big(\frac{1}{\alpha}Q(s(t),\{a^d_{m,j} ,\overline{a}_m(t)\};\theta_Q)\big)\big]\Big)\notag\\ 
&=\frac{1}{\alpha}\sum_{n=1}^N\pi_m(a^d_{m,n}|s(t);\theta_d)\Big[\alpha\log\pi_m(a^d_{m,n}|s(t);\theta_d)-Q\big(s(t),\{a^d_{m,n} ,\overline{a}_m(t)\};\theta_Q\big)\Big] \notag\\ 
&\quad +\sum_{n=1}^N\pi_m(a^d_{m,n}|s(t);\theta_d)\log\big[\sum_{j=1}^{N}\exp\big(\frac{1}{\alpha}Q(s(t),\{a^d_{m,j} ,\overline{a}_m(t)\};\theta_Q)\big)\big]\notag\\
&=\frac{1}{\alpha}\sum_{n=1}^N J^\pi_{m,n}(\theta_d) + \log\big[\sum_{j=1}^{N}\exp\big(\frac{1}{\alpha}Q(s(t),\{a^d_{m,j} ,\overline{a}_m(t)\};\theta_Q)\big)\big] \\[0.3cm]
J^{\pi}(\theta_{d}) &= J^{\pi}(\theta_{d})=\frac{1}{M}\sum_{m = 1}^{M} J^{\pi}_m(\theta_{d})\notag\\
&=\frac{1}{\alpha M}\sum_{m = 1}^{M}\sum_{n = 1}^{N}J^{\pi}_{m,n}(\theta_{d}) + \log\big[\sum_{j=1}^{N}\exp\big(\frac{1}{\alpha}Q(s(t),\{a^d_{m,j} ,\overline{a}_m(t)\};\theta_Q)\big)\big]\label{provedpolicyloss}
\end{align}
Which the second object in $J^{\pi}(\theta_{d})$ is a constant independent to the network parameters $\theta_{d}$, so it can be excepted from the policy loss function. Nevertheless, the adaptive temperature $\alpha$ before the summation in $J^{\pi}(\theta_{d})$ is unnecessary. To brevity, in our practical SDAC algorithm we use Eq.\eqref{policyloss} instead of \ref{provedpolicyloss}, which takes the same effect on the decomposed policy network.
\end{proof}
\clearpage

\subsection{KL divergence to square error loss function}\label{appc3}
In this section we focus on proving Eq.\eqref{supervisedequality}. The KL divergence loss function $J^{\pi}_m(\theta_{d})$ reduces to the variance $\sigma_{\pi_m}^2$ of $Q - Q_d$ at its minimum point, which $Q_d(s(t),\bigcdot \ ;\theta_d)- Q(s(t),\{\bigcdot \ ,\overline{a}_m(t)\} ;\theta_Q)=\delta$ for any discrete action in dimension $m$.
\begin{proof} 
For simplification, we assume an array $X$ to represent the error between $Q_d$ and $Q-\sigma$:
\begin{align}
X = \begin{bmatrix}
	x_1 \\[0.05cm]
	x_2 \\
	\vdots \\
	x_N
\end{bmatrix}
=\begin{bmatrix}
Q_d(s(t),a^d_{m,1} ;\theta_d)- Q(s(t),\{a^d_{m,1} ,\overline{a}_m(t)\} ;\theta_Q)-\delta \\[0.1cm]
Q_d(s(t),a^d_{m,2} ;\theta_d)- Q(s(t),\{a^d_{m,2} ,\overline{a}_m(t)\} ;\theta_Q)-\delta \\
\vdots\\
Q_d(s(t),a^d_{m,N} ;\theta_d)- Q(s(t),\{a^d_{m,N} ,\overline{a}_m(t)\} ;\theta_Q)-\delta
\end{bmatrix}
\end{align}
Which we have the KL divergence loss function$ \ J^{\pi}_m(\theta_{d})|_{X=O}=0$. Additionally, we use $Q_{m,n}$ to represent $Q(s(t),\{a^d_{m,n} ,\overline{a}_m(t)\} ;\theta_Q)$, and $Q^d_{m,n}$ to $Q_d(s(t),a^d_{m,n} ;\theta_d)$. With the Boltzmann policy in Eq.\eqref{boltzmann}; we can shift the KL divergence to the following object:
\begin{align}
J^{\pi}_m(\theta_{d}) 
&={\rm D_{KL}} \bigg(\frac{\exp \big(\frac{1}{\alpha} Q^d_{m, \bigcdot}\big)}{\sum_{j=1}^{N}\exp \big(\frac{1}{\alpha} Q^d_{m, j}\big)}\Bigg|\Bigg|
\frac{\exp\big(\frac{1}{\alpha}Q_{m, \bigcdot}\big)}{\sum_{j=1}^{N}\exp\big(\frac{1}{\alpha}Q_{m, j}\big)} \bigg)\notag\\	
&=\sum_{n=1}^N\pi_m(a^d_{m,n}|s(t);\theta_d)\Big[\frac{1}{\alpha}Q^d_{m, n} - \log\sum_{j=1}^{N} \exp \big(\frac{1}{\alpha} Q^d_{m, j}\big) -\frac{1}{\alpha}Q_{m, n} \notag\\
&\quad+ \log\sum_{j=1}^{N} \exp \big(\frac{1}{\alpha} Q_{m, j}\big)\Big]\notag\\
&=-\log\sum_{n=1}^{N} \exp \big(\frac{1}{\alpha} (x_n + Q_{m,n} + \sigma)\big)  + \log\sum_{n=1}^{N} \exp \big(\frac{1}{\alpha} Q_{m, n}\big)\notag\\
&\quad +\sum_{n=1}^N\pi_m(a^d_{m,n}|s(t);\theta_d)\big[\frac{1}{\alpha}(x_n+\sigma)\big].
\end{align}
Since $ \log\sum_{n=1}^{N} \exp \big(\frac{1}{\alpha} Q_{m, n}\big)$  and $\sigma$ are constants  independent to the parameters $\theta_d$, we can replace these objects by a constant $\xi$ in $J^{\pi}_m(\theta_{d})$ to ensure the loss function reaches $0$ at its minimum point: 
\begin{align}
	J^{\pi}_m(\theta_{d})
	&=\sum_{n=1}^N\frac{x_n}{\alpha}\bigcdot \pi_m(a^d_{m,n}|s(t);\theta_d) - \log\sum_{n=1}^{N} \exp \big(\frac{x_n + Q_{m,n}}{\alpha}\big)+\xi\notag\\
	&=\sum_{n=1}^N\frac{x_n}{\alpha}\bigcdot\frac{\exp \big(\frac{1}{\alpha} Q^d_{m, n}\big)}{\sum_{j=1}^{N}\exp \big(\frac{1}{\alpha} Q^d_{m, j}\big)}- \log\sum_{n=1}^{N} \exp \big(\frac{x_n + Q_{m,n}}{\alpha}\big)+\xi\notag\\[0.15cm]
	&=\dfrac{\sum_{n=1}^N x_n\exp \big(\dfrac{x_n + Q_{m, n}}{\alpha}\big)}{\alpha\sum_{n=1}^{N}\exp \big(\dfrac{x_n + Q_{m, n}}{\alpha}\big)}- \log\sum_{n=1}^{N} \exp \big(\frac{x_n + Q_{m,n}}{\alpha}\big)+\xi\label{xloss}
\end{align}
\clearpage
We marked \eqref{xloss} as $J^{\pi}_m(X)$, with its first-order partial deviation as follows:

\begin{align}\label{partial1}
	&\frac{\partial J^{\pi}_m(X)}{\partial X} = \\
	&\begin{bmatrix}
		\dfrac{x_1\exp \! \big(\dfrac{x_1\! + \!Q_{m, 1}}{\alpha}\big)\! \sum_{n =1}^N \exp \! \big(\dfrac{x_n \! +\! Q_{m, n}}{\alpha}\big)
		\!-\! \exp \! \big(\dfrac{x_1\! + \! Q_{m, 1}}{\alpha}\big)\! \sum_{n =1}^N x_n\exp \! \big(\dfrac{x_n \! +\! Q_{m, n}}{\alpha}\big)}
		{\Big[\alpha\sum_{n=1}^{N}\exp \big(\dfrac{x_n + Q_{m, n}}{\alpha}\big)\Big]^2} \\[0.05cm]
	\vdots \\[0.1cm]
		\dfrac{x_N\exp \!\big(\dfrac{x_N \! +\! Q_{m, N}}{\alpha}\big)\! \sum_{n =1}^N \exp \!\big(\dfrac{x_n \! +\! Q_{m, n}}{\alpha}\big)
		\!-\!\exp \!\big(\dfrac{x_N \!+\! Q_{m, N}}{\alpha}\big)\! \sum_{n =1}^N x_n\exp \!\big(\dfrac{x_n \! + \! Q_{m, n}}{\alpha}\big)}
		{\Big[\alpha\sum_{n=1}^{N}\exp \big(\dfrac{x_n +  Q_{m, n}}{\alpha}\big)\Big]^2}
	\end{bmatrix}\notag
\end{align}

Eq.\eqref{partial1} holds that  $\partial J^{\pi}_m(X) / \partial X = O$ when $X = O$. In order to explore the relationship between $\partial J^{\pi}_m(X)$ and the square error losses, We figured the second-order Taylor expansion of $J^{\pi}_m(X)$. The second-order partial deviation of $J^{\pi}_m(X)$ at $X=0$ can be described as the following object:
\begin{align}
	&\frac{\partial^2 J^{\pi}_m(X)}{\partial x_j^2} \bigg|_{X=0} =	
	\dfrac{\exp\big(\dfrac{Q_{m, j}}{\alpha}\big)\sum_{n=1, n\neq j}^N \exp  \big(\dfrac{ Q_{m, n}}{\alpha}\big)}	
	{\Big[\alpha\sum_{n=1}^{N}\exp \big(\dfrac{Q_{m, n}}{\alpha}\big)\Big]^2} \label{partial11}\\[0.3cm]
	&\frac{\partial^2 J^{\pi}_m(X)}{\partial x_j\partial x_k} \bigg|_{X=0} = 
	\dfrac{-\exp\big(\dfrac{Q_{m, j}}{\alpha}\big)\exp\big(\dfrac{Q_{m, k}}{\alpha}\big)}	
	{\Big[\alpha\sum_{n=1}^{N}\exp \big(\dfrac{Q_{m, n}}{\alpha}\big)\Big]^2}\label{partial12}	
\end{align} 
The 2-stage Taylor expansion of $J^{\pi}_m(X)$ at $X=O$ can be described by \eqref{partial1}, \eqref{partial11} and \eqref{partial12}:
\begin{align}\label{Taylor}
	&\lim\limits_{X \rightarrow O} \! J^{\pi}_m(X) \\[0.3cm]
	&= J^{\pi}_m(O)+ X^T\Big[\frac{\partial J^{\pi}_m(X)}{\partial X}\Big]_{\! X \! =O} 
	+ \frac{1}{2}X^T
	\begin{bmatrix}
		 \dfrac{\partial^2 J^{\pi}_m(X)}{\partial x_1^2}  & \cdots & \dfrac{\partial^2 J^{\pi}_m(X)}{\partial x_1\partial x_N} \\[0.05cm]
		\vdots & \ddots&\vdots \\[0.1cm]
		\dfrac{\partial^2 J^{\pi}_m(X)}{\partial x_1\partial x_N} &\cdots &\dfrac{\partial^2 J^{\pi}_m(X)}{\partial x_1^2} 
	\end{bmatrix}_{ \! \! X \! = O \!}
	X+ o(X^2) \notag\\[0.5cm]
	&= \dfrac{1}{2}X^T \!\!
	\begin{bmatrix}
		\dfrac{
		\exp \! \big(  \! \dfrac{ Q_{m, 1} } {\alpha} \! \big)
		\! \sum_{n=1, n\neq j}^N \!
		\exp \! \big(  \! \dfrac{ Q_{m, n} } {\alpha} \! \big)}	
		{\Big[\alpha\sum_{n=1}^{N}\exp \big(\dfrac{Q_{m, n}}{\alpha}\big)\Big]^2}\!\!\!
		& \cdots 
		& \dfrac{
		-\exp\big(  \! \dfrac{ Q_{m, 1} } {\alpha} \! \big)
		\exp\big(  \! \dfrac{ Q_{m, N} } {\alpha} \! \big)}	
		{\Big[\alpha\sum_{n=1}^{N}\exp \big(\dfrac{Q_{m, n}}{\alpha}\big)\Big]^2}
		 \\[0.05cm]
		\vdots & \ddots&\vdots \\[0.1cm]
		\dfrac{
		-\exp\big(  \! \dfrac{ Q_{m, N} } {\alpha} \! \big)
		\exp\big(  \! \dfrac{ Q_{m, 1} } {\alpha} \! \big)}	
		{\Big[\alpha\sum_{n=1}^{N}\exp \big(\dfrac{Q_{m, n}}{\alpha}\big)\Big]^2} 
		&\cdots 
		&\!\!\dfrac{
		\exp \! \big(  \! \dfrac{ Q_{m, 1} } {\alpha} \! \big)
		\! \sum_{n=1, n\neq j}^N \! 
		\exp \! \big(  \! \dfrac{ Q_{m, n} } {\alpha} \! \big)}	
		{\Big[\alpha\sum_{n=1}^{N}\exp \big(\dfrac{Q_{m, n}}{\alpha}\big)\Big]^2}
	\end{bmatrix}
	\!\! X \! + o(X^2) \notag
\end{align}
The Hessian matrix in Eq.\eqref{Taylor} can be split in to two objects, which corresponding to $\mathbb{E}_{\pi_m}[X^2]$ and $\mathbb{E}_{\pi_m}(X)^2$:
\begin{align}
	&\lim\limits_{X \rightarrow O}   J^{\pi}_m(X) \notag\\	
	&= \dfrac{1}{2\Big[ \alpha   \sum_{  n=1   }^{N}  \exp    \big(   \dfrac{Q_{  m,  n}}{\alpha}   \big)\Big]^2}\bigg(
	X^T   
	\begin{bmatrix}
		\exp\big(    \dfrac{ Q_{m, 1} } {\alpha}   \big)           & &  \\[0.05cm]
		& \ddots&\\[0.05cm]
		&  &          \exp\big(    \dfrac{ Q_{m, N} } {\alpha}   \big)
	\end{bmatrix}    X \bigcdot \sum_{n=1}^{N}  \exp    \big(    \dfrac{ Q_{m, n} } {\alpha}   \big) \notag\\[0.5cm]
	&\quad - X^T   
	\begin{bmatrix}
		\exp\big(    \dfrac{ Q_{m, 1} } {\alpha}   \big)^2  &\cdots & 
		\exp\big(    \dfrac{ Q_{m, 1} } {\alpha}   \big)
		\exp\big(    \dfrac{ Q_{m, N} } {\alpha}   \big) \\[0.05cm]
		\vdots& \ddots& \vdots \\[0.1cm]
		\exp\big(    \dfrac{ Q_{m, N} } {\alpha}   \big)
		\exp\big(    \dfrac{ Q_{m, 1} } {\alpha}   \big)
	& \cdots & \exp\big(    \dfrac{ Q_{m, 1} } {\alpha}   \big)^2
	\end{bmatrix}  X\bigg)+ o(X^2)\notag\\[0.3cm]
	&=\dfrac{1}{2\alpha^2}\bigg(
	\dfrac{\sum_{n=1}^N x_n^2\exp\big(    \dfrac{ Q_{m, n} } {\alpha}   \big)}
	{\sum_{  n=1   }^{N}  \exp    \big(   \dfrac{Q_{  m,  n}}{\alpha}   \big)}
	- \dfrac{\sum_{  n=1   }^{N} \sum_{  j=1   }^{N}x_n x_j 
	\exp\big(    \dfrac{ Q_{m, n} } {\alpha}   \big)
	\exp\big(    \dfrac{ Q_{m, j} } {\alpha}   \big)}
	{\Big[    \sum_{  n=1   }^{N}  \exp    \big(   \dfrac{Q_{  m,  n}}{\alpha}   			\big)\Big]^2}\bigg)+ o(X^2)\notag\\[0.2cm]
	&=\dfrac{1}{2\alpha^2}\! \bigg(\!\sum_{n=1}^N x_n^2\frac{
		\exp\big(  \! \dfrac{ Q_{m, n} } {\alpha} \! \big)}
	{{\sum_{  j=1 \! }^{N}  \exp  \! \big( \! \dfrac{Q_{\! m,  j}}{\alpha} \! \big)}}
	\!- \! \sum_{  n=1 \! }^{N}x_n
	\dfrac{\exp\big(  \! \dfrac{ Q_{m, n} } {\alpha} \! \big)}
	{\sum_{  k=1 \! }^{N}  \exp  \! \big( \! \dfrac{Q_{\! m,  k}}{\alpha} \! \big)}
	\sum_{  j=1 \! }^{N}x_j
	\dfrac{\exp\big(  \! \dfrac{ Q_{m, j} } {\alpha} \! \big)}
	{\sum_{  k=1 \! }^{N}  \exp  \! \big( \! \dfrac{Q_{\! m,  k}}{\alpha} \! \big)}\!\bigg) 
	\! + \! o(X^2)
	\notag\\[0.2cm]
	&=\dfrac{1}{2\alpha^2}\bigg(\sum_{n=1}^N\pi_m(a^d_{m,n}|s(t);\theta_d)\bigcdot x_n^2	
	-\Big[ \sum_{n=1}^N\pi_m(a^d_{m,n}|s(t);\theta_d)\bigcdot x_n\Big]^2\bigg) + o(X^2)\notag\\[0.2cm]
	&=\dfrac{1}{2\alpha^2}\mathbb{E}_{\pi_m}\bigg[\Big(Q_d\big(s(t),\bigcdot \ ;\theta_d\big)- Q\big(s(t),\{\bigcdot,\overline{a}_m(t)\};\theta_Q\big)-\delta\Big)^2\bigg] \notag\\[0.1cm]
	&\quad -  \bigg(\mathbb{E}_{\pi_m}\Big[Q_d\big(s(t),\bigcdot \ ;\theta_d\big)- Q\big(s(t),\{\bigcdot,\overline{a}_m(t)\};\theta_Q\big)-\delta\Big]\bigg)^2+ o(X^2)\notag\\[0.2cm]
	&=\dfrac{1}{2\alpha^2}\sigma_{\pi_m}^2\Big[Q_d\big(s(t),\bigcdot \ ;\theta_d\big)- Q\big(s(t),\{\bigcdot,\overline{a}_m(t)\};\theta_Q\big)-\delta\Big]+ o(X^2)\notag\\[0.2cm]
	&=\dfrac{1}{2\alpha^2}\sigma_{\pi_m}^2\Big[Q_d\big(s(t),\bigcdot \ ;\theta_d\big)- Q\big(s(t),\{\bigcdot,\overline{a}_m(t)\};\theta_Q\big)\Big]+ o(X^2)\label{supereqproof}
\end{align}
To further simplifications, we can discard $\dfrac{1}{2\alpha^2}$ in \eqref{supereqproof}, which takes little effects on the policy optimization process. Hence, \eqref{supereqproof} leads to our conclusions in Eq.\eqref{supervisedequality}.
\end{proof}

\clearpage
\renewcommand{\thesection}{Appendix D}
\section{Continuous Control Tasks based on MuJoCo and Box2d Environments}
\label{appd}
\begin{figure}[H]
	\centering
	\includegraphics[width=4in]{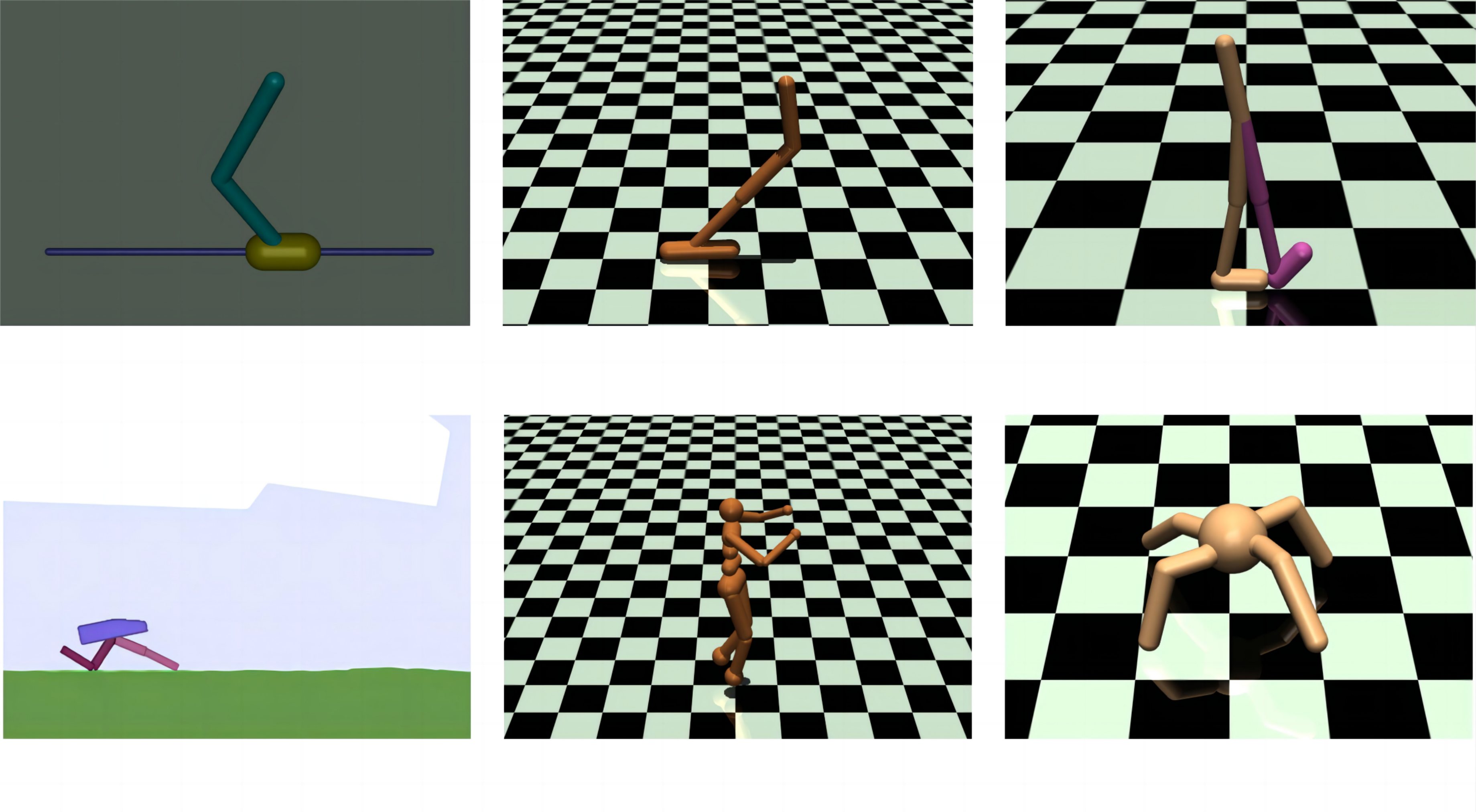}
\caption{OpenAI Gym baseline continuous control tasks. Top: InvertedDoublePendulum-v2, Hopper-v2, Walker2d-v2. Bottom: BipedalWalker-v3, Humanoid-v2, Ant-v2.}
	\label{img11}
\end{figure}

The continuous control tasks used to evaluate our algorithms are based on the MuJoCo physics engine, Box2d Engine the OpenAI Gym benchmark suite. These tasks are summarized in Figure \ref{img11} and feature high-dimensional continuous state spaces. For instance, Humanoid-v2 has $376$ continuous states, and Ant-v2 has $111$ continuous states. Some tasks also involve high-dimensional continuous action spaces to assess the effectiveness of our algorithms in handling such scenarios. We also include tasks with simpler environments, such as InvertedDoublePendulum-v2, which has only one continuous action dimension. Besides the 3D continuous control tasks, our baseline BipedalWalker-v3 task is based on 2D environment, which tested the generalization of our methods.

Evaluations in these environments have shown that our proposed algorithms perform well in a variety of settings. The attributes of all six continuous control tasks are listed in Table \ref{t4}. It's worth mentioned that we adjust the failure reward in BipedalWalker-v3 from -100 to -1, which large punishments would leads to invalid training results in a series of benchmark environments. 

\begin{table}[H]
	\caption{Attributes of the baseline environments}
	\label{t4}
	\begin{center}
		\begin{tabular}{p{130pt}l|lll} 
			
			\toprule 
			Environment & & State dimension & Action dimension &with termination \\
			
			\midrule 
			InvertedDoublePendulum-v2   & & 11    &  1&True\notag \\[0.1cm]
			Hopper-v2   & & 11    &  3&True\notag \\[0.1cm] 
			Walker2d-v2   & & 17    &  6&True\notag \\[0.1cm]
			BipedalWalker-v3   & & 24    &  4&False\notag \\[0.1cm]
			Humanoid-v2   & & 376    &  17&True\notag \\[0.1cm]
			Ant-v2   & & 111    &  8&True\notag \\

			\bottomrule 
		\end{tabular}
	\end{center}
\end{table}

\clearpage

\renewcommand{\thesection}{Appendix E}
\section{Ablation Studies}\label{appe}

\subsection{Studies on  the accuracy of action discretization}\label{appe1}

In our algorithms with decomposed discrete policies, it is necessary to discretize each continuous action space into $N$ discrete actions before the training process. However, the choice of $N$ can influence the algorithm's training efficiency. In this section, we conducted experiments to evaluate our algorithm SDAC on several baseline environments using different values of $N$, specifically $N=10, 20$, and $50$.

The experimental results demonstrate that higher values of $N$ tend to lead to more effective performances. Conversely, when a small value of $N$ is used, indicating low accuracy in action discretization, the algorithm may fail to update an effective policy. Both SDAC and SDCQ, our proposed algorithms, can handle relatively high values of $N$ without experiencing training failures. However, if an excessively large value of $N$ is incorporated, the policy network with $M\times N$ output neurons may become difficult to train. Moreover, higher values of $N$ correspond to increased computational complexity.

\begin{figure}[H]
	\centering
	\includegraphics[width=5.5in]{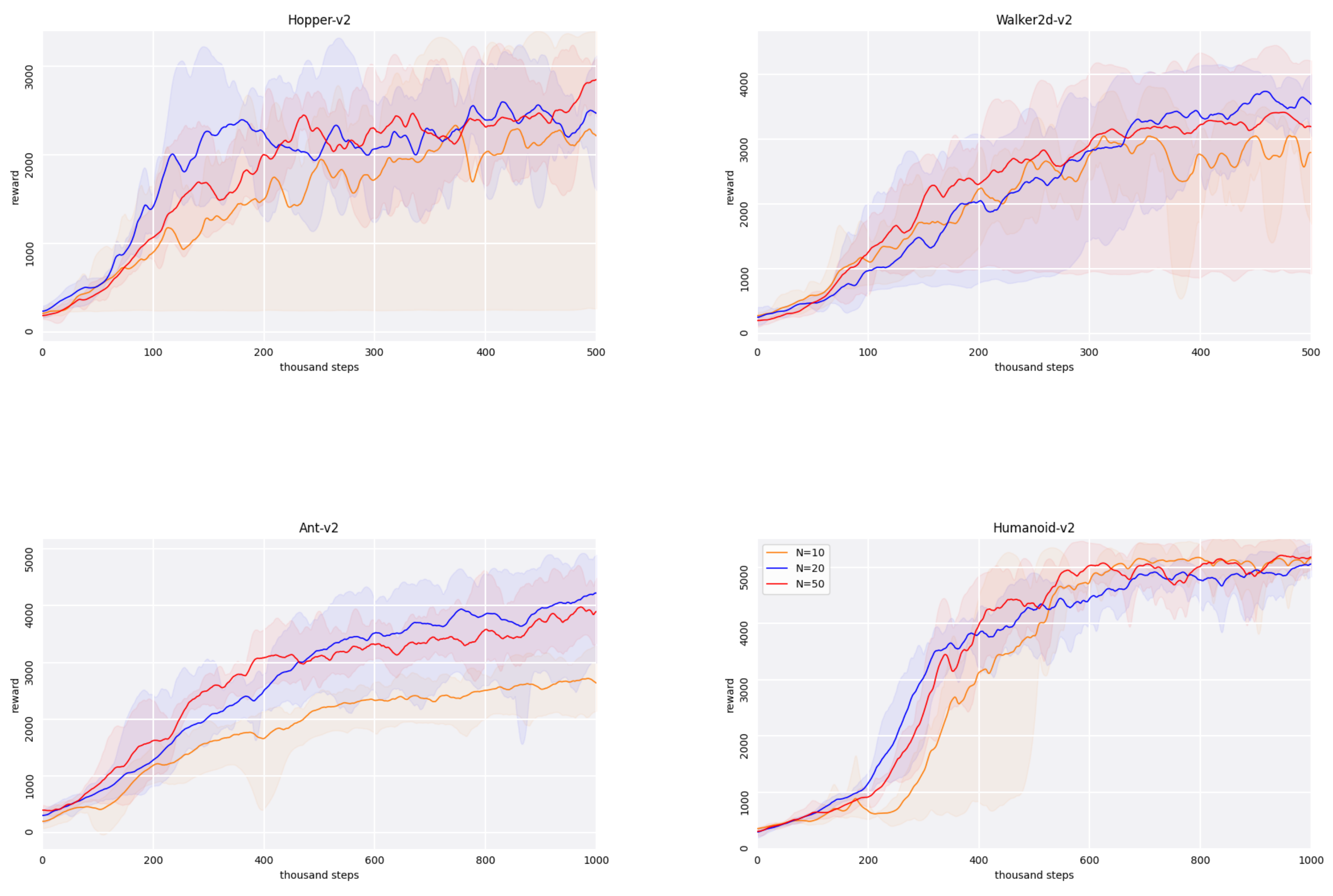}
\caption{Performance results for different action discretization levels $N$ in SDAC.}
	\label{img12}
\end{figure}

Throughout our evaluations, we typically set $N$ to 20. The evaluation results for different action discretization levels in SDAC are shown in Figure \ref{img12}.

\clearpage

\subsection{Studies on SDAC's target entropy}\label{appe2}

The target entropy $\hat{\mathcal{H}}$ is a critical hyperparameter for soft RL algorithms with adaptive temperature adjustment, which $\hat{\mathcal{H}}$ determines the exploration rate of the algorithm, similar to the deviation of Gaussian exploration noises in TD3\cite{fujimoto2018addressing}. In the continuous soft RL algorithm SAC\cite{haarnoja2018soft2}, they set their target entropy to $-\dim(\mathcal{A})$, corresponding to  $\hat{\mathcal{H}}=-1$ in our SDAC and SDCQ.

\begin{figure}[H]
	\centering
	\includegraphics[width=5.5in]{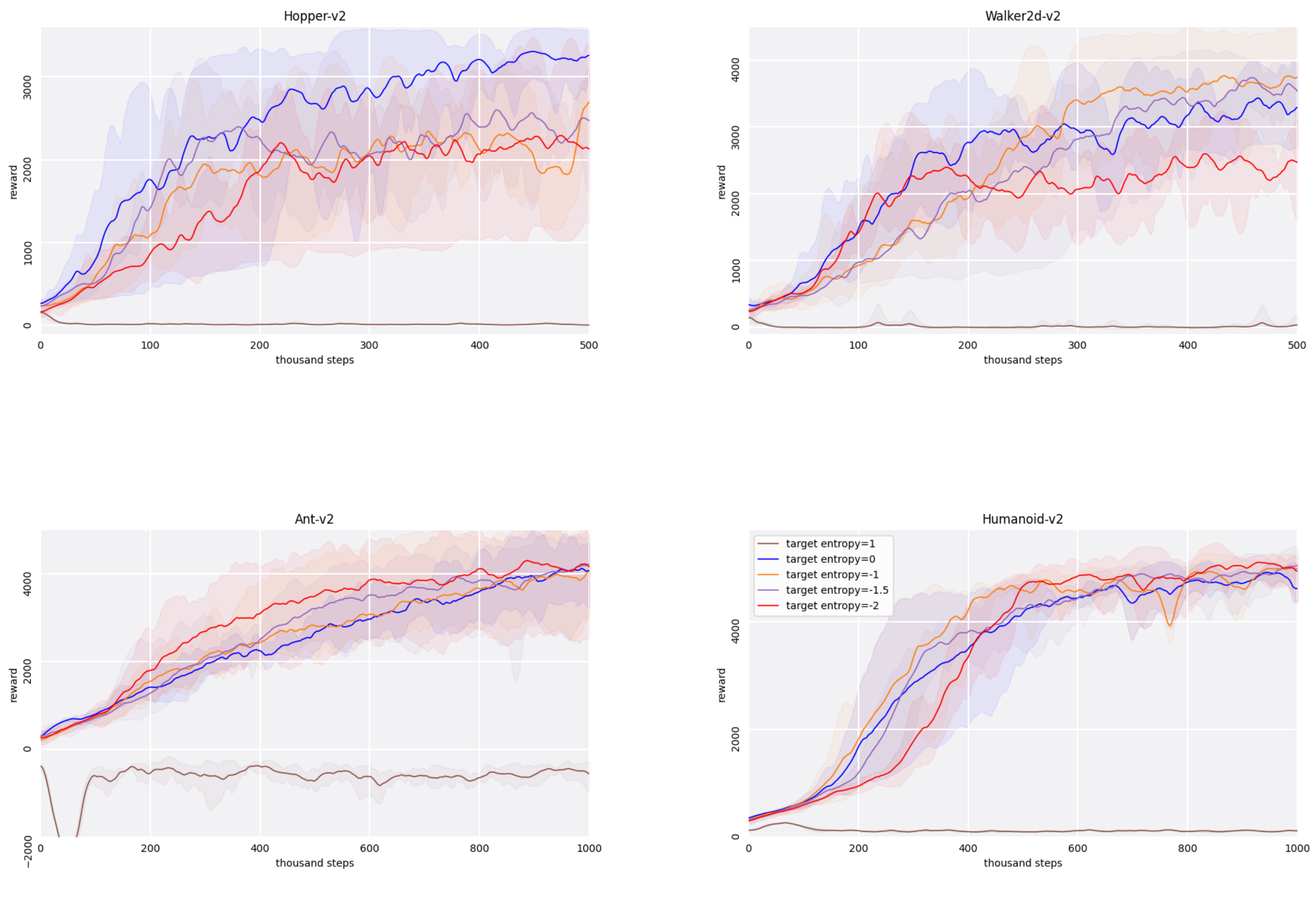}
\caption{Performance results for different target entropy levels $\hat{\mathcal{H}}$ in SDAC.}

	\label{SDAC target entropy}
\end{figure}

In our evaluations, we have tested different target entropy in SDAC, from -2 to 1. As it demonstrated in figure.\ref{SDAC target entropy}, it's fair for SDAC with an target entropy from -2 to 0, but $\hat{\mathcal{H}}=1$ leads to exceeded exploration rate, which cause training failures in all of the three baseline environments. Results also show that the environments have different preferences on exploration rate, which Hopper-v2 prefer larger $\hat{\mathcal{H}}$, but a smaller $\hat{\mathcal{H}}$ preforms better on Walker2d-v2 or Ant-v2.

\clearpage

\subsection{Studies on SDCQ's target entropy}\label{appe3}

The target entropy $\hat{\mathcal{H}}=1$ take different effects on SDCQ compared with SDAC. In all of the four benchmark environments, SDCQ prefers a larger $\hat{\mathcal{H}}=0$, which leads to higher exploration rate and better performances. 

Besides, $\hat{\mathcal{H}}=-1$ would cause training failures in both SDAC and SDCQ.

\begin{figure}[H]
	\centering
	\includegraphics[width=5.5in]{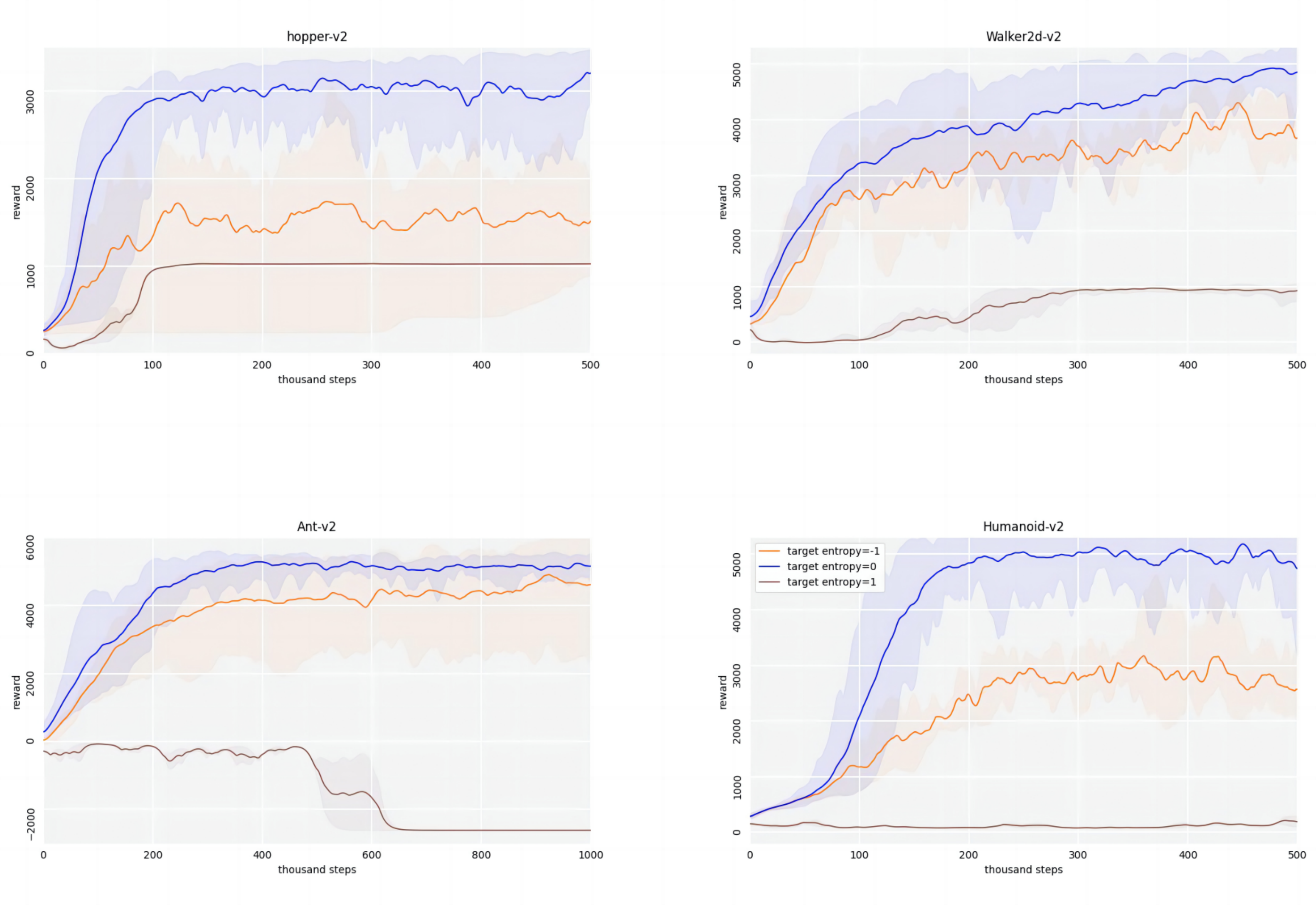}
	\caption{Performance results for different target entropy levels $\hat{\mathcal{H}}$ in SDCQ.}
	
	\label{SDCQ target entropy}
\end{figure}

\clearpage

\subsection{Ablation studies on multi-step TD}\label{appe4}
\begin{figure}[H]
	\centering
	\includegraphics[width=5.5in]{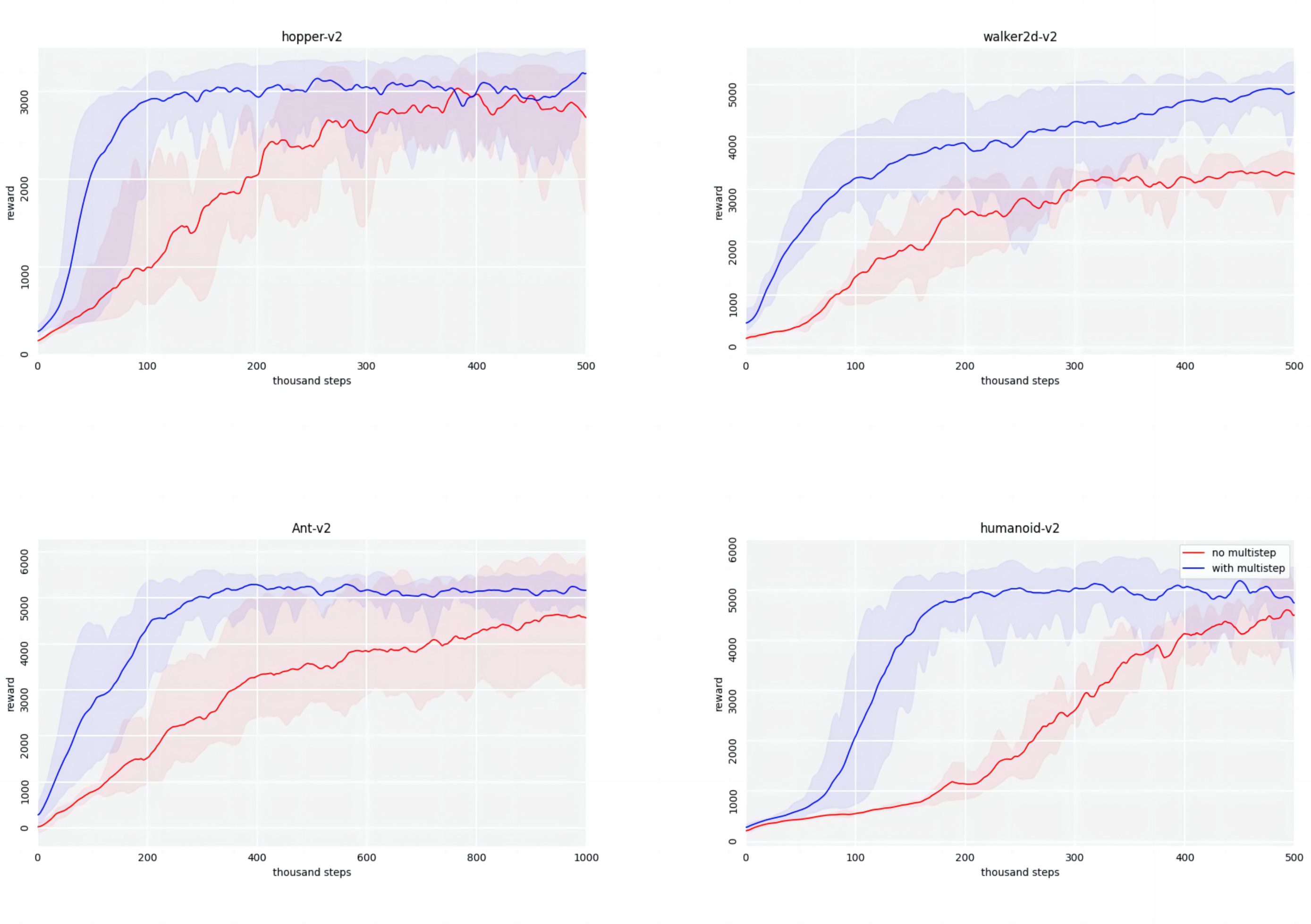}
	\caption{Ablation studies on the Multi-step TD of SDCQ, averaged over 5 random seeds.}
	\label{ablationmultistep}
\end{figure}

Since multi-step TD is effective to the value-based DQN algorithm, it can also significantly improve the performance of our discrete SDCQ algorithm on the continuous control tasks. We presents the ablation study of the multi-step object in our SDCQ algorithm in figure.\ref{ablationmultistep}. It's approved by the experimental results that the multi-step object presents stable and effective enhancements on our SDCQ algorithm, which can significantly improve the training efficiency on all of the four MuJoCo benchmark environments.
\clearpage

\subsection{Ablation studies on the normalized importance sampling}\label{appe5}

\begin{figure}[H]
	\centering
	\includegraphics[width=5.5in]{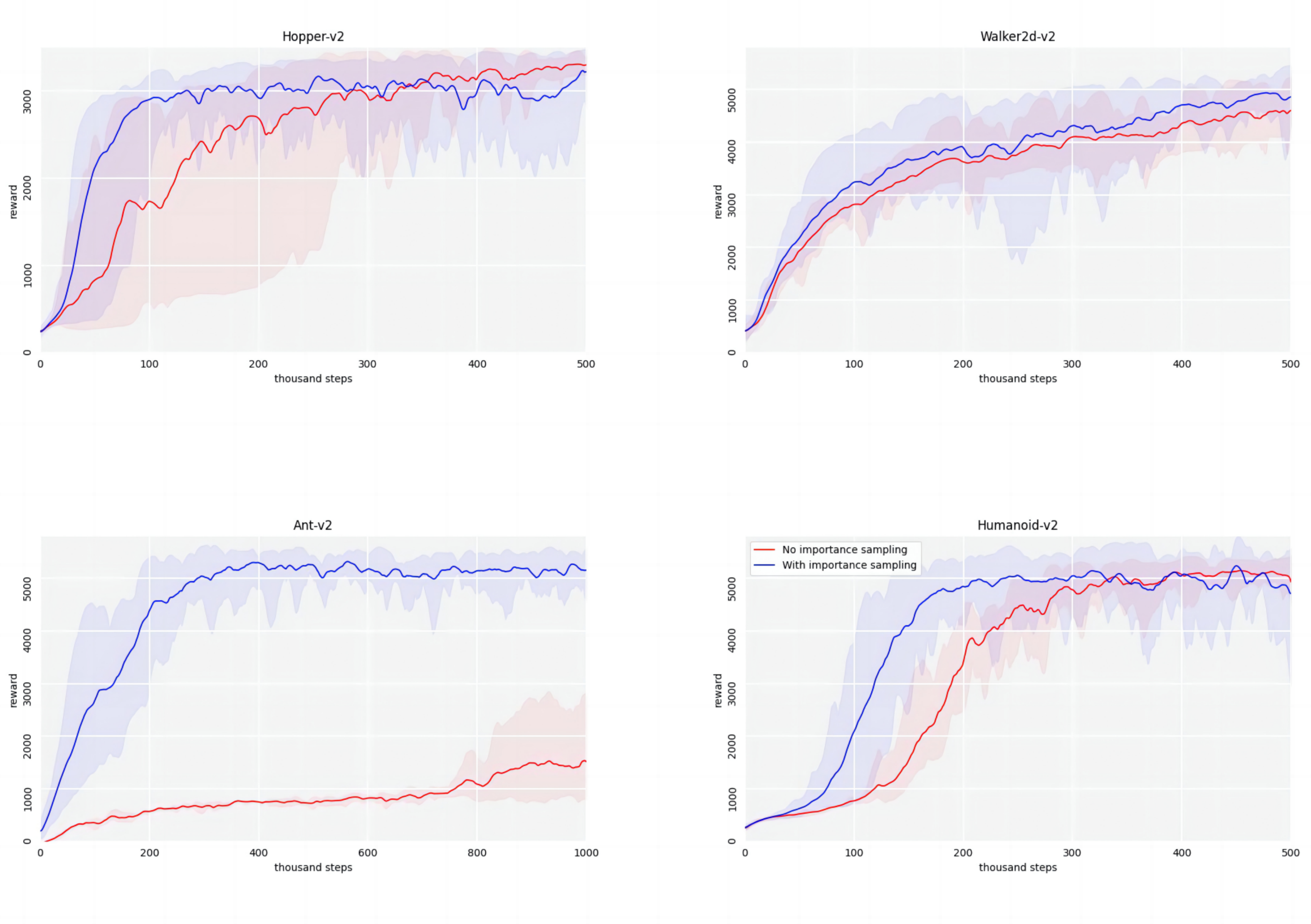}
\caption{Ablation studies on the normalized importance sampling of SDCQ, averaged over 5 random seeds.}
	\label{ablationimportance}
\end{figure}

In this section, we provide the ablation studies for our normalized importance sampling in SDCQ. The main purpose of the object is to correct the biases presented in off-policy multi-step TD, and improve the stability of our practical SDCQ algorithm. Figure.\ref{ablationimportance} shows the comparisons between multi-step SDCQ with or without normalized importance sampling, which indicates the effectiveness of the object on all of the four benchmark environments. Some tasks, like Ant-v2, may face training failures without the object.

\clearpage

\subsection{Ablation studies on the target adaptive temperature}\label{appe6}
\begin{figure}[H]
	\centering
	\includegraphics[width=5.5in]{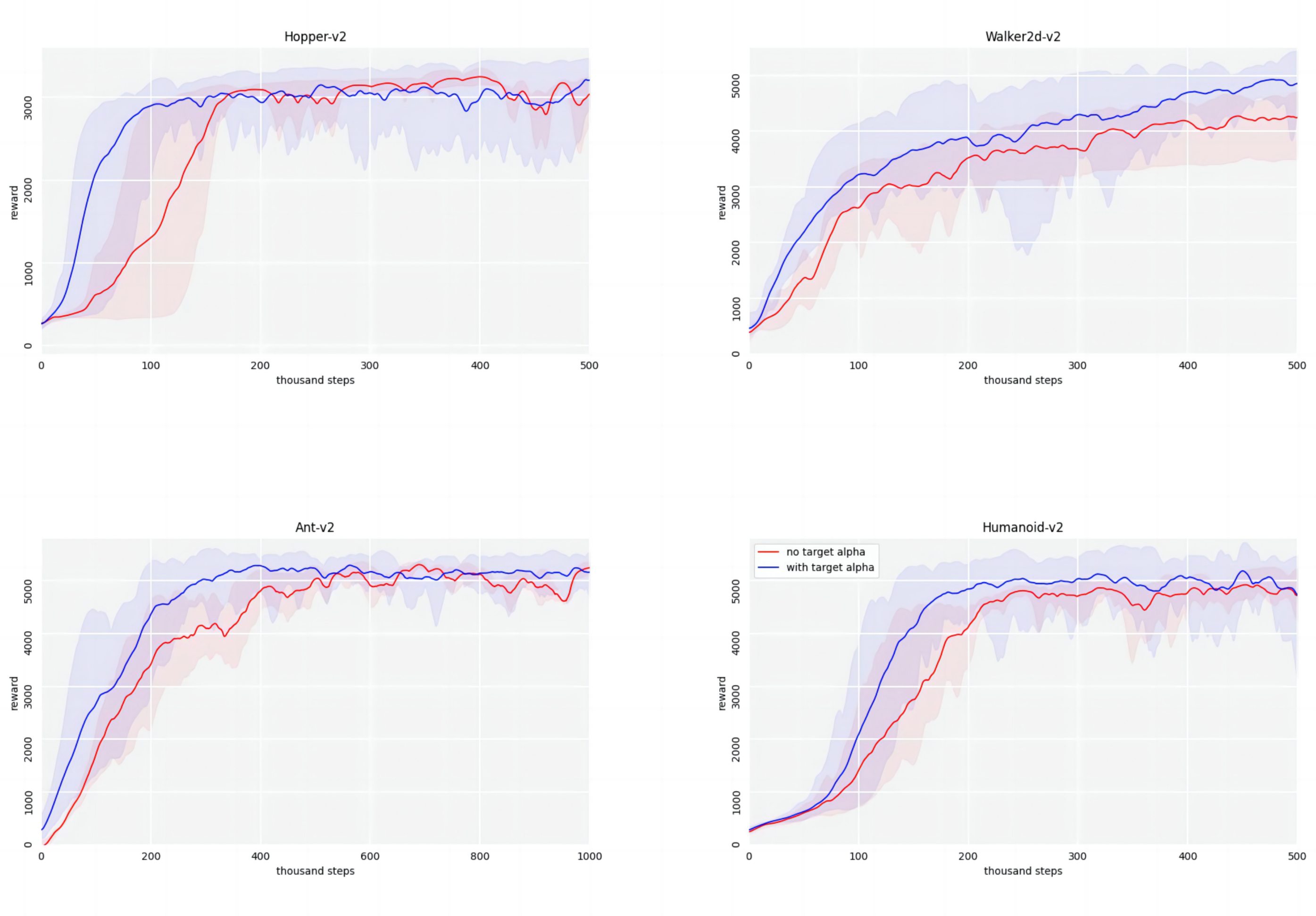}
	\caption{Ablation studies on the target adaptive temperature $\alpha'$ of SDCQ with multi-step TD, averaged over 5 random seeds.}
	\label{ablationtemperature}
\end{figure}

In our SDCQ algorithm, the adaptive temperature can strictly influence the distribution of the exploration policy, and the target policy when sampling $\widetilde{a}(t+1)$. During our evaluations, we found the adaptive temperature $\alpha$ varies rapidly, which may disturb the value iterations of the continuous critic network. Figure.\ref{ablationtemperature} shows the effect of $alpha'$ on SDCQ, which the object can stabilize training process effectively. 

Additionally, the $\alpha'$ plays a critical role in the simplified SDCQ algorithm without multi-step TD. But in our practical SDCQ, the multi-step object may take effects stabilizing the adaptive temperature, which made the algorithm less sensitive to $\alpha'$.

\clearpage

\end{document}